\documentclass[11pt]{article}

\usepackage[preprint]{acl}

\usepackage{times}
\usepackage{latexsym}

\usepackage[T1]{fontenc}
\usepackage[utf8]{inputenc}

\usepackage{microtype}

\usepackage{inconsolata}

\usepackage{graphicx}

\usepackage{url}
\usepackage{booktabs}
\usepackage{amsfonts}
\usepackage{nicefrac}
\usepackage{xcolor}
\usepackage{xspace}
\usepackage{subcaption}
\usepackage{multirow}
\usepackage{adjustbox}
\usepackage{colortbl}
\usepackage{array}
\usepackage{amsmath}
\usepackage{amssymb}
\usepackage{pifont}
\usepackage{caption}
\usepackage{tikz}
\usetikzlibrary{arrows.meta, positioning}
\usepackage{tcolorbox}
\tcbuselibrary{skins}
\usepackage{enumitem}
\usepackage{wrapfig}
\usepackage{float}
\usepackage{hyperref}

\newcommand{\rw}{\textsc{RouterWorld}\xspace}

\newcommand{\gc}[1]{{\color{teal}\scriptsize #1}}
\newcommand{\rc}[1]{{\color{red}\scriptsize #1}}

\definecolor{cardheader}{HTML}{D97706}
\definecolor{cardbg}{HTML}{FFFBEB}
\definecolor{cardborder}{HTML}{F59E0B}
\definecolor{tagmax}{HTML}{DC2626}
\definecolor{tagvery}{HTML}{EA580C}
\definecolor{tagcon}{HTML}{2563EB}
\definecolor{tagnorm}{HTML}{7C3AED}
\definecolor{textmuted}{HTML}{6B7280}
\definecolor{promptsys}{HTML}{1E40AF}
\definecolor{promptsysbg}{HTML}{EFF6FF}
\definecolor{promptuser}{HTML}{047857}
\definecolor{promptuserbg}{HTML}{ECFDF5}
\definecolor{promptout}{HTML}{9333EA}
\definecolor{promptoutbg}{HTML}{F5F3FF}
\definecolor{promptvar}{HTML}{B45309}
\definecolor{promptframe}{HTML}{CBD5E1}
\definecolor{catRollout}{HTML}{1A7F37}
\definecolor{catJudge}{HTML}{8250DF}
\definecolor{catAnnot}{HTML}{BF8700}
\definecolor{catRouter}{HTML}{1F6FEB}
\definecolor{catAgent}{HTML}{CF222E}

\title{RouterWorld: Scaling Conversational AI with\\Latency-Aware LLM Routing}

\author{Jinyang Li$^{1,\dagger}$ \hspace{0.9em} Denis Emelin$^{2}$ \hspace{0.9em} Simone Tedeschi$^{2}$ \hspace{0.9em} Weiwei Cheng$^{2}$ \hspace{0.9em} Cristina Morariu$^{2}$\\ {\bf Haytham Assem}$^{2}$ \hspace{0.4em} {\bf Jens Lehmann}$^{2}$ \hspace{0.4em} {\bf Emine Yilmaz}$^{2,3}$\thanks{Emine Yilmaz holds concurrent appointments as an Amazon Scholar and as a Professor of Computer Science at University College London. This paper describes work performed at Amazon.} \hspace{0.4em} {\bf Angus Addlesee}$^{2}$\\\\
$^1$The University of Hong Kong \hspace{0.3em}
$^2$Amazon \hspace{0.3em}\\
$^3$Centre for Artificial Intelligence, University College London\\\\
$^\dagger$Work done while at Amazon\\
}

\begin{document}
\maketitle

\begin{abstract}
Conversational AI systems built on Large Language Models (LLMs) require routing mechanisms that balance answer quality and response latency at scale. However, existing LLM routing benchmarks assume full observability of all candidate model outcomes, under-emphasize latency as a first-class metric, and fragment evaluation across heterogeneous protocols misaligned with real-world settings. We introduce \textbf{\rw{}}, a comprehensive benchmark for LLM routing in conversational GenAI. \rw{} unifies \textbf{371k} single-turn responses and \textbf{76k} multi-turn agentic trajectories, adopts natural user interactions with a standardized LLM-as-a-judge protocol, and decouples efficiency from transient pricing via hardware-grounded latency annotations. Crucially, \rw{} formalizes routing under partial observability through two setups: \textbf{Cold-Start}, which restricts learning to sparse, static logs, and \textbf{Voyager}, which models continuous data streams where routing decisions shape future training distributions. Evaluating 13 routing baselines, we find that most routers over-emphasize accuracy while ignoring latency under partial observability. Code and benchmark data are publicly available.
\end{abstract}

\section{Introduction}

Conversational AI systems powered by large language models (LLMs) increasingly operate at massive scale, serving millions of user queries per minute across chat assistants \cite{team2023gemini, llama2023llama, chatarena, amazon2025alexa}, search interfaces \cite{chen2025generative, eval-search}, and interactive agents \cite{react2023yao, wu2024autogen, toollm2024qin}. In these settings, every query is user-facing and latency-sensitive. Rather than relying on a single monolithic model, modern conversational systems deploy heterogeneous pools of LLMs that vary widely in capability, specialization, and inference latency. Intelligent routing, which dynamically selects the most appropriate model for each query, has thus emerged as a central mechanism for scaling Generative AI while balancing response quality and cost.

\begin{figure*}[t]
    \centering
    \includegraphics[width=\textwidth]{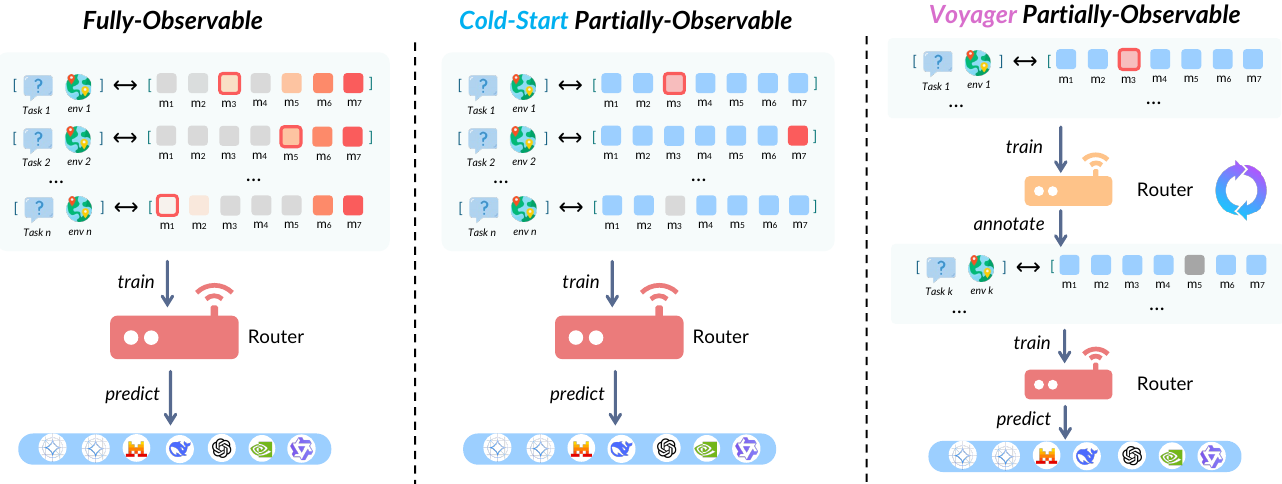}
    \caption{\textbf{Routing observability taxonomy.}
    (\textit{Left}) \textbf{Fully-Observable} routing, where feedback from all candidate models is available.
    (\textit{Center}) \textbf{Cold-Start Partially-Observable} routing, where the router observes outcomes only for selected models.
    (\textit{Right}) \textbf{Voyager Partially-Observable} routing, where routing decisions affect future data collection and training. $m_i$ refers to $i^{th}$ candidate model.
    \includegraphics[height=0.9em]{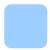} denotes models not selected by the router,
    \includegraphics[height=0.9em]{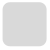} denotes incorrect outputs,
    while vivid colors indicate correct predictions.
    Color intensity reflects inference latency (darker indicates higher latency),
    and \includegraphics[height=0.9em]{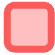} dark borders mark optimal models that are both correct and fastest.}
    \label{fig:observability}
\end{figure*}

Recent work has explored various routing strategies, including supervised classifiers trained on offline logs \cite{routellm,hybridllm,graphrouter}, bandit-based methods that optimize routing policies via online interaction \cite{pilot,barp}, and cascade architectures that progressively escalate queries to stronger models \cite{frugalgpt}. To facilitate comparison across methods, benchmarks have been proposed \cite{routerbench,llmrouterbench,routereval}.

Despite this progress, existing evaluation methodologies for LLM routing remain misaligned with real-world deployment constraints. A central limitation is the widespread assumption of full observability: as illustrated in Figure~\ref{fig:observability} (Left), standard protocols assume the router has access to correctness and efficiency outcomes of \emph{all} candidate models per query. This diverges sharply from deployed systems operating at million-scale throughput, where running all candidates per request is prohibitively expensive and incompatible with rapid iteration cycles~\cite{chapelle2011empirical,joachims2017unbiased}. In practice, outcomes of unchosen models remain counterfactual and unavailable for learning. Existing benchmarks also frequently use monetary costs as efficiency proxies \cite{dean2013tail}, neglecting latency. Because natural dialogue relies on $\sim$200ms inter-turn gaps \cite{stivers2009universals}, multi-second delays break conversational fluidity, so evaluation metrics must ground efficiency in stable, hardware-level characteristics.

To bridge these gaps, we introduce \textbf{\rw}, a comprehensive benchmark that aligns LLM routing research with the operational realities of conversational AI. \rw{} consolidates 53k single-turn core tasks and 17k multi-turn agentic interactions across 7 open-weight LLMs (2B--235B parameters, covering coding, math, reasoning, and chit-chat specialists), for a total of \textbf{371k} single-turn responses and \textbf{76k} multi-turn trajectories once every candidate model is run on every task. To reflect deployed input--output characteristics, we replace artifact-heavy task prompts with natural \textbf{Conversational-Form Queries (CF-Queries)}, unify heterogeneous metrics into a binary correctness signal via a standardized LLM-as-a-judge protocol aligned with human evaluation, and replace volatile commercial pricing with reproducible latency metrics obtained from the GenAI-Perf\footnote{\url{https://docs.nvidia.com/deeplearning/triton-inference-server}} toolkit on controlled hardware.

Crucially, \rw{} models the counterfactual nature of routing feedback through two partially observable evaluation paradigms (Figure~\ref{fig:observability}). The \emph{Cold-Start} setup simulates zero-shot initialization, training routers from static historical logs without access to unselected model outcomes. The \emph{Voyager} mode extends this to a dynamic, life-long learning setting in which routing decisions shape future training data, exposing the system to non-stationary traffic and testing its ability to mitigate self-reinforcing bias through exploration.

Benchmarking 13 routing strategies on \rw{} reveals a pronounced efficiency gap. To jointly measure routing quality and efficiency, we introduce \textbf{L-DCG}, a latency-aware variant of Discounted Cumulative Gain that weighs each correct selection by how close its latency is to the fastest correct model (formal definition in Section~\ref{sec:metrics}). Even under idealized full observability, the strongest baseline reaches only 39.6\% latency-penalized accuracy (Acc@1-L) compared to the 68.0\% empirical upper bound, capturing just 62\% of the optimal efficiency ceiling as measured by L-DCG, indicating substantial headroom in latency-aware routing. This gap widens substantially under more realistic partial observability: deprived of counterfactual feedback, routers usually over-emphasize task accuracy (Acc@1) at the expense of latency, inflating accuracy by an average of 24.7\% while overall efficiency deteriorates sharply. In the most extreme case, \textsc{Avenger}~\cite{avenger}, a cluster-voting router, collapses into a degenerate \emph{Route-to-Largest} policy \cite{lai2026routing}, achieving 89.1\% Acc@1 but suffering a 79\% reduction in optimal efficiency. These feedback-induced failures persist in the Voyager setting, where standard routers show little improvement across streaming interaction, while bandit-based methods with explicit exploration, such as \textsc{BaRP}~\cite{barp}, continue to improve, highlighting observation diversity. These findings motivate the next generation of router design.

Further, we provide analyses to further clarify routing behavior. We examine long-horizon agentic tasks, where policy collapse emerges through cumulative latency effects rather than single-step decisions (Section \ref{sec:rw_agent}), and study warm-up initialization, finding that most routers function as policy distillers rather than independent optimizers (Section \ref{sec:warm_up}). By open-sourcing \rw-Core and \rw-Agentic\footnote{\href{https://dataverse.harvard.edu/previewurl.xhtml?token=17dc0733-b88a-4411-9fab-360f89418a44}{RouterWorld-Core} and \href{https://dataverse.harvard.edu/previewurl.xhtml?token=2de555ec-1a26-48d2-8bc1-7904b5d0750e}{RouterWorld-Agentic} datasets.}, we provide a scalable, deployment-aligned environment for diagnosing these failures and advancing robust, latency-aware routing for conversational AI.

\section{Related Work}
\label{sec:related_work}

\paragraph{Non-Predictive and Sequential Routing.}
Early LLM routing relied on non-predictive orchestration, where decisions are resolved through iterative execution rather than a single selection step. FrugalGPT~\cite{frugalgpt} formalizes this via the LLM cascade, escalating queries to progressively stronger models when a confidence threshold is not met. More recent work generalizes this into ensemble-like frameworks: AutoMix~\cite{automix} formulates routing as a POMDP with self-verification signals, and Router-R1~\cite{zhang2025routerr} instantiates the router as a reasoning LLM that plans multi-step invocations. While flexible, these approaches invoke multiple models per query, leading to compounded inference latency.

\paragraph{Predictive Routing and Online Model Selection.}
Predictive approaches formulate model selection as a single decision made prior to generation. HybridLLM~\cite{hybridllm} trains a BERT-based classifier to route between a small and a large model, while RouteLLM~\citep{routellm} and Carrot~\cite{tsiourvas2025causal} learn from preference or reward signals. GraphRouter~\cite{graphrouter} models query--model interactions via graph-based message passing, BaRP~\citep{barp} formulates routing as learning from bandit feedback using REINFORCE-style policy gradients, and Avenger~\citep{avenger} employs train-free cluster-based voting over embedding neighborhoods. IPR~\cite{feng-etal-2025-ipr} targets production settings with user-controlled quality--cost trade-offs, and PILOT~\cite{pilot} explores online contextual bandits with exploration--exploitation schedules tailored to routing.

\paragraph{Fully-Observable Routing.}
Despite this variety of methods, almost all predictive routers are designed and evaluated under \emph{full observability}: at training time, the learner has access to the outcome of every model on every query, reducing routing to supervised learning over complete labels. This assumption underlies RouteLLM~\citep{routellm}, HybridLLM~\cite{hybridllm}, GraphRouter~\cite{graphrouter}, Carrot~\cite{tsiourvas2025causal}, Avenger~\citep{avenger}, and the Best-of-$N$ style oracles commonly reported as upper bounds \cite{routerbench,routereval}. While convenient, full observability requires running every model per query, an operational pattern that does not scale to production-scale conversational deployments which preclude exhaustive parallel inference~\cite{chapelle2011empirical,joachims2017unbiased}. It also obscures feedback-loop dynamics (confirmation bias, policy collapse, self-reinforcement) that only emerge when the router's choices shape the data it subsequently learns from. A few methods acknowledge partial feedback, notably BaRP~\citep{barp}, but its evaluation still assumes rich offline supervision. Our benchmark makes partial observability a first-class evaluation regime and exposes how existing routers behave when counterfactual outcomes are not available.

\paragraph{Router Benchmarks.}
Routing methods are typically compared using benchmarks such as RouterBench~\cite{routerbench}, RouterArena~\cite{routerarena}, RouterEval~\cite{routereval}, and LLMRouterBench~\cite{llmrouterbench}, which release precomputed outcomes for many query--model pairs. These resources assume full observability and either omit latency or treat it as a coarse boundary constraint rather than a primary optimization objective. They also rely on raw benchmark prompts and monetary pricing as efficiency proxies, both of which introduce artifacts and drift that are ill-suited to conversational deployment. \rw{} instead (i) formalizes partial observability via the Cold-Start and Voyager setups, (ii) grounds efficiency in hardware-measured latency that is stable across pricing changes, (iii) adopts natural conversational-form queries, and (iv) spans both single-turn and long-horizon agentic tasks, offering a more deployment-aligned evaluation. 

\section{Problem Definition}
\subsection{LLM Routing}\label{sec:routing_general}

Let $\mathcal{M} = \{m_1, \dots, m_K\}$ denote a fixed set of candidate LLMs. A routing policy
\[
\pi : \mathcal{S} \rightarrow \mathcal{M}
\]
maps an observed state $s \in \mathcal{S}$ to a selected model $m \in \mathcal{M}$. The state $s$ represents the full or partial input available to the router. After selecting a model, the router observes a signal reflecting task performance and computational efficiency. In this work, we focus on latency-derived efficiency (Section \ref{sec:model_and_latency}) rather than monetary cost. The routing objective is to learn a policy $\pi$ that maximizes the \textit{expected utility} of the system under these performance and latency constraints.

\subsection{Fully-Observable LLM Routing} \label{sec:fully_observable}
In the fully-observable setting, the router has access to outcomes of all candidate models for each input during training, reducing routing to supervised learning over complete labels. This has been the dominant assumption in the LLM routing literature to date, underlying most predictive routers and benchmarks surveyed in Section~\ref{sec:related_work} (e.g., \citealp{hybridllm,routellm,graphrouter,tsiourvas2025causal,avenger,routerbench,routereval}). It is convenient for offline analysis but relies on counterfactual outcomes that are typically unavailable in practice without costly offline replay across all candidate models.

\subsection{Partially-Observable LLM Routing}
\label{sec:partial_observable}

We propose a more realistic scenario of partial observability, where the router observes feedback \textit{only} for the selected model. Formally, after choosing $m = \pi(s)$, the router observes $r(s, m)$, but the counterfactuals $\{r(s, m')\}_{m' \neq m}$ remain missing. The learner must contend with incomplete information. We study two modes of partial observability:

\paragraph{Cold-Start Routing.}

The Cold-Start setting represents the initialization phase of a routing system. The router must learn from a static historical log generated by a uniform random policy, where each task yields a single model outcome (response, latency, and correctness). This evaluates the router's ability to extract a viable policy from high-entropy, incomplete data in a single offline training round. We explore other warm-up policies in Section~\ref{sec:partially_analysis}.

\paragraph{Voyager Routing.}

In real-world deployments, routing decisions are made over streaming inputs, and the router's own policy determines which model outcomes are observed. \emph{Voyager} mode captures this: training proceeds in iterative stages over temporally ordered data chunks $\{C_1, \dots, C_T\}$. The process is initialized with $C_1$ (ground-truth feedback under an initial random policy); at stage $t$, the policy is trained on $\{C_1, \dots, C_t\}$ and used to route $C_{t+1}$, with checkpoint selection on a held-out development set. To keep evaluation reproducible without live model inference, we simulate deployment via a \textit{hidden fully-observable engine} derived from \rw: it stores outcomes for all candidates but reveals only the feedback corresponding to the router's selected action.

\section{Benchmark Construction}

\subsection{Dataset Selection Principles}
\label{sec:dataset_selection}

We construct \rw according to four guiding principles: (1) \textbf{Task Diversity and Balance}, spanning six major categories (e.g., math, code, reasoning) to prevent domain skew; (2) \textbf{Discriminative Power}, selecting examples where model performance varies to necessitate intelligent routing; (3) \textbf{Latency Awareness}, integrating real-world timing metrics \cite{orca, vllm} to evaluate trade-offs; and (4) \textbf{Reproducibility}, enforcing strict train-test separation and automatic metrics. Table~\ref{tab:benchmark_comparison} compares \rw with existing benchmarks. Unlike prior work, our benchmark uniquely combines partial observability, streaming evaluation, and agentic workflows; see Appendix~\ref{app:dataset_principles} for a full breakdown.

\begin{table*}[t]
\centering
\resizebox{\linewidth}{!}{%
\begin{tabular}{lrrrcccccc}
\toprule
\textbf{Benchmark} & \textbf{\#Instances} & \textbf{\#Models} & \textbf{\#Datasets} & \textbf{\#Base.} & \textbf{Part.\ Obs.} & \textbf{Stream.} & \textbf{Latency Ann.} & \textbf{Agentic Tasks} & \textbf{CF-Query} \\
\midrule
RouterBench~\cite{routerbench}        & 405K & 11 & 8  & 4  & \ding{55} & \ding{55} & \ding{55} & \ding{55} & \ding{55} \\
RouteLLM~\cite{routellm}           & ---  &  2 & 3  & 4  & \ding{55} & \ding{55} & \ding{55} & \ding{55} & \ding{55} \\
LLMRouterBench~\cite{llmrouterbench}     & 400K & 33 & 21 & 10 & \ding{55} & \ding{55} & \ding{55} & \ding{55} & \ding{55} \\
RouterEval~\cite{routereval}         & 200M & 8{,}500+ & 12 & 5  & \ding{55} & \ding{55} & \ding{55} & \ding{55} & \ding{55} \\
RouterArena~\cite{routerarena}        & 8.4K & 42 & 23 & 14 & \ding{55} & \ding{55} & \ding{55} & \ding{55} & \ding{55} \\
\midrule
\textsc{RouterWorld-Core}    & 53K & 7 & 12 & 17 & \textcolor{green!60!black}{\ding{51}} & \ding{55} & \textcolor{green!60!black}{\ding{51}} & \ding{55} & \textcolor{green!60!black}{\ding{51}} \\
\textsc{RouterWorld-Agentic} & 17K & 3 & 3  & 17 & \textcolor{green!60!black}{\ding{51}} & \textcolor{green!60!black}{\ding{51}} & \textcolor{green!60!black}{\ding{51}} & \textcolor{green!60!black}{\ding{51}} & \textcolor{green!60!black}{\ding{51}} \\
\bottomrule
\end{tabular}%
}
\caption{Comparison of LLM routing benchmarks. ``Part.\ Obs.'' = partially observable setting; ``Stream.'' = streaming rerouting; ``Latency Ann.'' = per-instance latency annotations; ``\#Base.'' = number of routing baselines evaluated; ``CF-Query'' = user query normalization as router input.}
\label{tab:benchmark_comparison}
\end{table*}

\subsection{RouterWorld Benchmark Construction}

\rw{} comprises two complementary subsets. \textbf{\rw-Core} contains single-turn, end-to-end tasks assessing a router's ability to select appropriate models for standalone queries of varying difficulty. It integrates 12 established benchmarks covering logical and commonsense reasoning~\citep{hellaswag, arc_challenge, fever, hotpotqa}, math~\citep{gsm8k, tabmwp, hendrycks} and tool use~\citep{gorilla, mcptoolbench++}, code generation~\citep{apps, birdsql}, data analysis~\citep{birdsql,tabmwp}, and open-ended chit-chat~\citep{personachat}. \textbf{\rw-Agentic} focuses on long-horizon, multi-turn agentic tasks that require environment interaction, covering computer use, household assistance, and e-commerce workflows. All benchmarks are standardized into a unified routing evaluation framework; full preprocessing and standardization procedures are in Appendix~\ref{app:dataset_details}.

\begin{figure}[t]
  \centering
  \includegraphics[width=\linewidth]{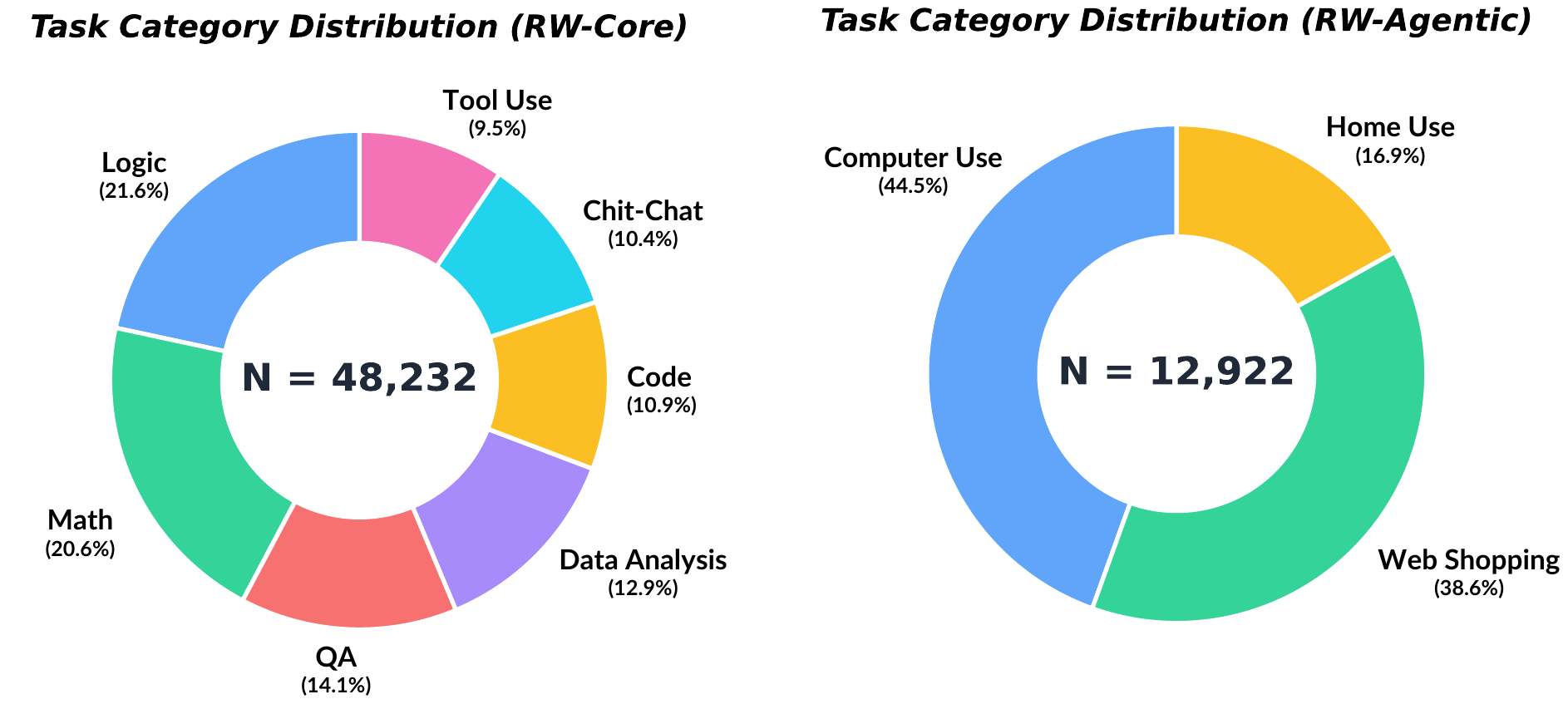}
  \caption{Task category distribution in \textsc{Router-Core} and \textsc{Router-Agentic}.}
  \label{fig:task-distribution}
\end{figure}

\paragraph{\rw Characteristics.}
Figure~\ref{fig:task-distribution} and Table~\ref{tab:data-characteristics} summarize the statistics. \textsc{\rw-Core} contains 48{,}232 single-turn instances across seven task categories and twelve datasets, with substantial variation in query difficulty and latency that creates meaningful routing opportunities. \textsc{\rw-Agentic} comprises 12{,}740 multi-turn interactions in three agentic domains and exhibits higher selectivity and longer latency tails. Analysis of task distributions, difficulty, model specialization, and oracle behavior is in Appendix~\ref{app:dataset_stats_core}.

\subsection{Annotation}

\paragraph{Model Selection \& Latency.} \label{sec:model_and_latency}
\label{sec:model_selection}

We select 7 popular open-weight models (2B--235B parameters), creating a heterogeneous landscape of capability and inference cost, where latencies range from $\sim$200ms to over 7s. We model E2E latency for query-model pairs $(q, m)$ using empirically measured Time-To-First-Token ($\text{TTFT}_m$) and Inter-Token Latency ($\text{ITL}_m$):
\begin{equation}
    \text{latency}(q, m) = \text{TTFT}_m + (N_{q,m} - 1) \times \text{ITL}_m,
\end{equation}
where $N_{q,m}$ is the number of tokens generated. For multi-step agentic trajectories, latency is accumulated across all generation steps. Since raw response and trajectory tokens are released for each task, the framework can be easily adapted to monetary cost. Full profiling workflow and model specifications are in Appendices~\ref{app:latency-profiling} and~\ref{app:model_details}.

\paragraph{Dataset Evaluation.}
Source datasets employ heterogeneous evaluation metrics, and individual models often require distinct prompting and post-processing pipelines. To streamline the setup and comparison while mitigating biases introduced by dataset-specific metrics and prompt engineering, we adopt an \emph{LLM-as-a-Judge} framework that produces a standardized binary judgment (\textit{correct} or \textit{incorrect}) for each model output via a strong LLM, i.e. \texttt{Claude-4.5-Sonnet}. This binary signal is intended to approximate coarse-grained user feedback in interactive systems, where users typically either adopt a satisfactory response or reject it to ask a follow-up query, rather than providing dense or calibrated rewards \cite{amershi2019guidelines,ouyang2022training}. We validate this protocol through a blind expert annotation study on 150 (query, model, response) triples: six PhD-level annotators achieve high inter-annotator agreement (Fleiss' $\kappa{=}0.84$, Krippendorff's $\alpha{=}0.87$), and the LLM judge matches the 6-expert majority-vote consensus at $\kappa{=}0.84$ (F1${=}0.93$, 92\% agreement), approaching the human ceiling and consistent with prior LLM-judge evaluations~\citep{zheng2023judging}. Full validation details are provided in Appendix~\ref{app:judge-validation}.

\paragraph{Conversational-Form Query Generation.}

Real user inputs to voice assistants are typically brief and goal-oriented rather than carefully structured prompts. Further, since benchmarks have been publicly accessible for years, downstream LLMs are susceptible to data contamination and may have memorized surface-level queries and prompt formats during pre-training~\cite{magar2022data,golchin2024time,sainz2023nlp}, artificially inflating performance and introducing spurious signals that mislead routers. We therefore apply a multi-stage transformation pipeline: original prompts establish task specifications; a strong external LLM (\texttt{DeepSeek-V3.1}, excluded from the routing pool) rewrites each as a natural user query; rewritten queries are validated against the original specs, with semantically drifted ones filtered out~\cite{li2025swesql, zheng2023judge, wang2023selfconsistency, maddan2023selfrefine}.

\subsection{Metrics}
\label{sec:metrics}

\paragraph{Accuracy@1 and Accuracy@1-L (Optimal).} We report two top-1 accuracy metrics for routing. \textit{Accuracy@1} measures whether the model selected by the router correctly answered the query. This metric captures the router's ability to select any correct model, independent of efficiency.

\textit{Accuracy@1-L} is a stricter, latency-aware variant: it equals $1$ if the router selects the fastest correct model for the query (i.e., the minimum-latency member of $\mathcal{C}_q$) and $0$ otherwise, including when the selected model is incorrect. This metric corresponds to the optimal routing decision under latency constraints and serves as a reference for evaluating latency-sensitive routing performance.

\paragraph{DCG and L-DCG.}

To evaluate rank-aware routing quality beyond top-1 accuracy, we adopt Discounted Cumulative Gain (DCG). In our setting, relevance is binary: a model is relevant if and only if it produces a correct answer.

For a query $q$, let $\mathcal{C}_q$ denote the set of correct models, ordered by ascending latency. If the router selects model $m^*$ with rank $r$ in this ordering, the DCG score is defined as
\begin{equation}
    \mathrm{DCG}(q) = \frac{\mathbf{1}\!\left[m^* \in \mathcal{C}_q\right]}{\log_2(r + 1)} .
\end{equation}

While DCG captures sensitivity to rank among correct models, it does not reflect the magnitude of latency differences. To incorporate efficiency, we define a latency-aware variant, \textbf{L-DCG}:
\begin{equation}
    \mathrm{L\text{-}DCG}(q) =
    \mathbf{1}\!\left[m^* \in \mathcal{C}_q\right] \cdot
    \frac{\ell_{\min}}{\ell_{m^*}} \cdot \frac{1}{\log_2(r + 1)},
\end{equation}
where $\ell_{\min}$ is the latency of the fastest correct model and $\ell_{m^*}$ is the latency of the selected model; the score is $0$ when the selected model is incorrect. The ratio $\ell_{\min}/\ell_{m^*} \in (0,1]$ continuously penalizes correct but suboptimal latency choices, reflecting efficiency loss relative to the optimal decision. All metrics are reported as averages over the test set.

\begin{table}[t]
\centering
\footnotesize
\setlength{\tabcolsep}{3pt}
\renewcommand{\arraystretch}{1.1}
\resizebox{\linewidth}{!}{%
\begin{tabular}{l c c}
\toprule
\textbf{Statistic} & \textbf{\textsc{Core}} & \textbf{\textsc{Agentic}} \\
\midrule
\textbf{Total Instances} & 48,232 & 12,740 \\
\quad \# Training        & 40,265 & 11,292 \\
\quad \# Testing          & 7,967  & 1,448  \\
\addlinespace[2pt]
\textbf{\# Task Categories} & 7   & 3      \\
\textbf{\# Datasets}        & 12  & 3      \\
\textbf{\# Models}           & 7   & 6      \\
\addlinespace[2pt]
Prompt len.\ (mean/max)      & 2.1k/31k & 1.1k/6.3k \\
Query len.\ (mean)           & 110      & 111       \\
Avg \# correct models        & 4.19     & 2.95      \\
\bottomrule
\end{tabular}%
}
\caption{Data characteristics of \rw.}
\label{tab:data-characteristics}
\end{table}

\begin{figure*}[h]
    \centering
    \includegraphics[width=\textwidth]{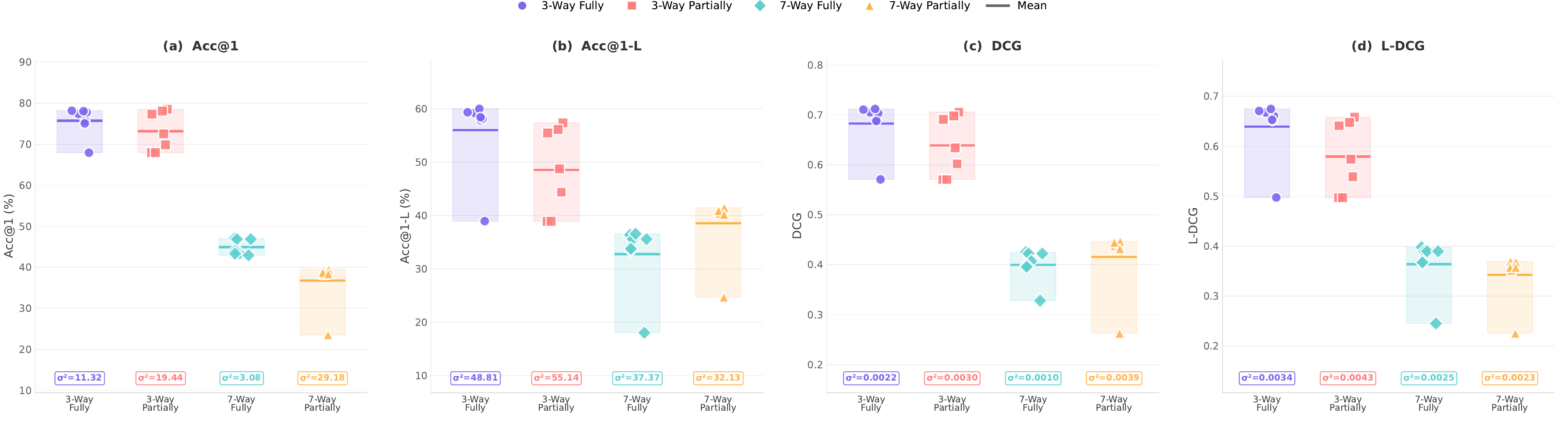}
    \caption{Performance compression in agentic routing. Each dot is a non-collapsed routing baseline. Under both 3-model and 6-model pools, methods cluster tightly with low variance across all metrics. Boxes show interquartile range; $\sigma^2$ values shown below.}
    \label{fig:agent_compression}
\end{figure*}

\section{Experiments}

\subsection{Experimental Setup}
\label{sec:exp_setup}

\paragraph{Baselines.}

We benchmark 13 popular routing algorithms across categories.
(1) \textbf{Naive routers} provide reference bounds, including \emph{Random} selection,
\emph{Always-Route-To-Largest} (always picking the largest candidate model, regardless of query) and \emph{Always-Route-To-Smallest} (always picking the smallest), as well as two oracle
upper bounds: \emph{Oracle (Perfect)}, which selects the optimal downstream model with
full knowledge of all candidate outcomes, and \emph{Oracle (Best-of-$N$)}, which, for each
query, picks the best downstream model among the $N$ models that at least one evaluated
router selects on that query, representing an empirical (rather than theoretical) upper
bound attainable by any router in our suite.
(2) \textbf{Representation-based routers} learn a query representation and route based
on similarity, classification, or learned utility in embedding space. This category
covers classical methods operating on frozen embeddings (\emph{KNN}, \emph{MLP}),
end-to-end representation fine-tuning (\emph{Emb-FT}), and more structured or hybrid
architectures such as \emph{Hybrid-LLM}~\cite{hybridllm}, \emph{BaRP}~\cite{barp},
\emph{GraphRouter}~\cite{graphrouter}, as well as \emph{Avenger}~\cite{avenger}.
(3) \textbf{Generative routers} use autoregressive language models to directly select a
target model via prompting or supervised fine-tuning on routing data.
Detailed architectures and training configurations for all baselines are provided in
Appendix~\ref{app:baselines}.

\paragraph{Observability Settings.}
We evaluate all methods under two regimes: (1) \textbf{Fully Observable}, where the training set contains ground-truth correctness and latency for every candidate model; and (2) \textbf{Partially Observable}, where each query is annotated with the outcome of only \emph{one} random model.

\paragraph{Implementation details.}
All experiments are conducted on $8\times$ NVIDIA H200 GPUs. Since most routing algorithms require a query encoder, we standardize on \texttt{Qwen3-Embedding-0.6B} (1024-dim, mean-pooled, max sequence length 512) across all methods for a fair comparison. For methods that require architecture-specific encoders (such as Hybrid-LLM which relies on DeBERTa's \texttt{[CLS]} token representation) we retain their original encoder as prescribed by the respective papers.

\subsection{Main Results}
\begin{table*}[t]
\centering
\label{tab:router_comparison}
\renewcommand{\arraystretch}{1.3}
\resizebox{1.0\textwidth}{!}{%
\begin{tabular}{l l c c c c c c c c}
\hline
& & \multicolumn{4}{c}{\textbf{Fully-Observable}} & \multicolumn{4}{c}{\textbf{Partially-Observable}} \\
\cline{3-6} \cline{7-10}
\textbf{Method} & \textbf{Base Model} &
\textbf{Acc@1 (\%)} & \textbf{Acc@1-L (\%)} & \textbf{DCG} & \textbf{L-DCG} &
\textbf{Acc@1 (\%)} & \textbf{Acc@1-L (\%)} & \textbf{DCG} & \textbf{L-DCG} \\
\hline

\rowcolor{gray!15} \multicolumn{10}{c}{\textit{Naive Baselines}} \\

\rowcolor{blue!15}
Oracle (Perfect) & -- &
100.00 & 100.00 & 1.0000 & 1.0000 &
100.00 & 100.00 & 1.0000 & 1.0000 \\

\rowcolor{blue!6}
Oracle (Best-of-8) & -- &
80.04 & 67.97 & 0.7521 & 0.7140 &
96.81 & 54.51 & 0.7901 & 0.6462 \\

Random & -- &
47.63 & 12.80 & 0.2995 & 0.1784 &
47.63 & 12.80 & 0.2995 & 0.1784 \\

Route-To-Largest & -- &
89.11 & 9.10 & 0.4731 & 0.1494 &
89.11 & 9.10 & 0.4731 & 0.1494 \\

Route-To-Smallest & -- &
33.16 & 26.55 & 0.3054 & 0.2942 &
33.16 & 26.55 & 0.3054 & 0.2942 \\

\rowcolor{gray!15} \multicolumn{10}{c}{\textit{Generative AI Routers}} \\

Zero-Shot Prompt & Qwen-2.5-0.5B-ins &
\textbf{78.65} & 7.38 & 0.3979 & 0.1211 &
78.65 & 7.38 & 0.3979 & 0.1211 \\

Gen-SFT & Qwen-2.5-0.5B-ins &
55.11 & \underline{37.55} & \underline{0.4767} & \underline{0.4282} &
81.15 & \underline{16.91} & \underline{0.4935} & \underline{0.2507} \\

Zero-Shot Prompt & Qwen-2.5-1.5B-ins &
43.89 & 10.79 & 0.2912 & 0.1953 &
43.89 & 10.79 & 0.2912 & 0.1953 \\

Gen-SFT & Qwen-2.5-1.5B-ins &
54.17 & \textbf{39.61} & 0.4817 & \textbf{0.4439} &
\underline{84.55} & 15.87 & \textbf{0.4959} & 0.2268 \\

\rowcolor{gray!15} \multicolumn{10}{c}{\textit{Representation-based Routers}} \\

KNN & Qwen3-Embedding-0.6B &
52.25 & 34.35 & 0.4456 & 0.3969 &
71.82 & 14.86 & 0.4317 & 0.2215 \\

MLP & Qwen3-Embedding-0.6B &
51.99 & \underline{38.20} & 0.4631 & 0.4295 &
83.56 & 14.59 & 0.4822 & 0.2158 \\

Emb-FT & Modern-base-0.1B &
53.26 & 33.41 & 0.4490 & 0.4041 &
75.10 & 15.28 & 0.4533 & 0.2342 \\

Emb-FT & Qwen3-Embedding-0.6B &
55.93 & 37.87 & \underline{0.4833} & \underline{0.4357} &
78.57 & 16.10 & 0.4702 & 0.2288 \\

Hybrid-LLM & DeBERTa Large &
47.92 & 35.95 & 0.4309 & 0.4039 &
66.93 & \textbf{19.52} & 0.4352 & \textbf{0.2661} \\

Graph-Router & Qwen3-Embedding-0.6B &
38.17 & 13.64 & 0.2674 & 0.1994 &
58.87 & 13.48 & 0.3590 & 0.1892 \\

BARP & Qwen3-Embedding-0.6B &
\underline{76.58} & 21.53 & \textbf{0.4966} & 0.2978 &
\textbf{88.57} & 9.66 & 0.4748 & 0.1565 \\

Avenger & Qwen3-Embedding-0.6B &
45.86 & 32.43 & 0.4026 & 0.3709 &
\cellcolor{orange!18}89.11 &
\cellcolor{orange!18}9.10 &
\cellcolor{orange!18}0.4731 &
\cellcolor{orange!18}0.1494 \\

\hline
\end{tabular}%
}
\caption{Performance on \rw-Core under fully- and partially-observable settings. Oracle (Perfect) and Oracle (Best-of-8) denote theoretical and empirical upper bounds, respectively. Orange cells indicate collapsed routing that always selects a single model and are excluded from best/second-best comparisons. All pairwise improvements of the best-performing method over the second-best baseline are statistically significant ($p<0.05$).}
\label{tab:main_result}
\end{table*}
\paragraph{\rw is Challenging.}

\rw poses a substantial challenge for existing routers (Table~\ref{tab:main_result}). Under full observability, the empirical upper bound \emph{Oracle (Best-of-8)} achieves 67.97\% Acc@1-L / 0.7140 L-DCG, whereas the strongest non-oracle baseline \emph{Gen-SFT} (Qwen-2.5-1.5B-ins) reaches only 39.61\% Acc@1-L / 0.4439 L-DCG, leaving a large gap in latency-aware routing. Effective routing requires context-aware decision-making rather than coarse model attributes, as \emph{Route-To-Largest} attains 89.11\% Acc@1 but just 9.10\% Acc@1-L


Transitioning to partial observability drives a drastic shift: on average, Acc@1 inflates by 24.7 pp while L-DCG drops by over 0.16. Without cross-model comparisons, routers cannot determine whether a faster model would have also been correct, and become risk-averse (favoring accuracy at the cost of latency). In the extreme, Avenger collapses into Route-To-Largest (89.11\% Acc@1, 0.1494 L-DCG), losing any ability to navigate efficiency trade-offs, a fundamental challenge that \rw explicitly captures.

\paragraph{Training Corpus is Important for Routers.}
To illustrate the necessity of domain-specific training data, we probe Generative AI Routers. Zero-shot prompting with explicit latency profiles severely underperforms (Qwen-2.5-0.5B-Instruct: 7.38\% Acc@1-L, 0.1211 L-DCG), while SFT on the \rw corpus drives Acc@1-L to 37.55\% and L-DCG to 0.4282.
This shows that routing cannot be reduced to a linear relationship between model size and query difficulty: in $\sim$27\% of cases, lightweight models like \texttt{Gemma-2-2b} exhibit verbose generation patterns that yield higher latency than larger models. LLM routing is a complex, non-linear decision-making process that requires internalizing empirical interaction logs.

\subsection{Routing in Agentic Tasks} \label{sec:rw_agent}

We further evaluate routing baselines on \rw-Agentic, where models interact with environments over multiple steps across three benchmarks with new 3-model and 6-model candidate pools (14B--235B). Figure~\ref{fig:agent_compression} (see Appendix~\ref{app:additional-figures}) shows a striking phenomenon: routing baselines exhibit severe performance compression. In both pools, non-collapsed methods cluster with low variance ($\sigma^2=0.81$ for Acc@1-L in the 3-model fully-observable setting; $\sigma^2=4.35$ in the 6-model setting; vs.\ $\sigma^2=29.5$ in \rw-Core), and compression intensifies under partial observability, where multiple routers collapse into degenerate near-single-model policies. We attribute this to a factor overlooked by current routing paradigms: the impact of planning steps. High-capability models often resolve complex tasks in far fewer interaction steps than weaker but faster models, neutralizing or even inverting the conventional accuracy--latency trade-off and leaving no meaningful Pareto frontier to exploit. Per-benchmark breakdowns and planning--latency analysis are in Appendix~\ref{app:agentic_results}.

\subsection{Analysis of Partially-Observable Routing}
\label{sec:partially_analysis}
\label{para:warmup}

We further analyze partially-observable routing: Voyager learning, and warm-up initialization.
\paragraph{Cold-Start v.s. Voyager}
The Voyager setting incrementally accumulates streaming data over five rounds. Surprisingly, this yields only marginal improvements for most methods (Figure~\ref{fig:voyager}, see Appendix~\ref{app:additional-figures}): KNN and MLP L-DCG remains effectively flat, while only \emph{BaRP} shows a noticeable gain ($0.294 \rightarrow 0.320$), consistent with its REINFORCE-based formulation. Supervised methods suffer from confirmation bias under partial observability, reinforcing their initial blind spots by training on their own past decisions; policy-gradient methods instead use observed feedback as an unbiased reward estimator and naturally explore alternatives. Collecting more data under the same biased policy is thus insufficient to mitigate distribution shift: observation \emph{diversity} matters more than volume.

\paragraph{Warm-Up Initialization.} \label{sec:warm_up}

To study how initial policy quality influences downstream learning, we simulate three warm-up oracles of increasing sophistication: Qwen-1.5B-Instruct, Claude, and Claude augmented with RAG retrieval from the dev set. Each rolls out selections to generate warm-up logs, on which we then train routers under cold-start, partially-observable conditions. As Figure~\ref{fig:warmup-wave} (see Appendix~\ref{app:additional-figures}) shows, downstream routers correlate near-perfectly with the metric distribution of each warm-up policy: no method surpasses the oracle's own L-DCG, and the ordering of Acc@1, Acc@1-L, DCG, and L-DCG is preserved. Current predictive routing thus largely distills the logging oracle's biases rather than discovering an independent policy, motivating exploration--exploitation strategies that actively seek diverse observations.

\section{Conclusion}
We introduced \rw{}, a benchmark that unifies 371k single-turn responses and 76k multi-turn agentic trajectories to evaluate LLM routing under realistic deployment constraints. RouterWorld addresses key gaps in existing benchmarks by formalizing partial observability through Cold-Start and Voyager settings, grounding efficiency in hardware-measured latency rather than volatile pricing, and adopting natural conversational-form queries. Evaluating 13 routing baselines reveals large headroom for latency-aware routing even under idealized conditions. Our analysis highlights the critical importance of partial observability as a design constraint. By open-sourcing our benchmark and evaluation infrastructure, RouterWorld serves as a foundation for advancing efficiency-aware LLM routing in scalable conversational AI.


\section*{Limitations}

\paragraph{Multi-modal Tasks.}
\rw{} currently evaluates routing only among text-only LLMs. As foundation models increasingly accept vision and audio inputs, practical routing systems may need to decide both \emph{which} model and \emph{which} modality to invoke. Extending \rw{} to multimodal candidate pools requires new datasets and evaluation protocols that capture heterogeneous latency and cost profiles across modalities, which we leave for future work.

\paragraph{Closed-source Models.}
\rw{} focuses on open-weight LLMs to enable reproducible, hardware-grounded latency measurement, which is central to our latency-aware routing framing. As a result, widely used closed-source models (e.g., Claude, GPT-4o) are not included in the candidate pool. Because we release raw responses, trajectory tokens, and generation metadata for every task--model pair, the benchmark can be extended to closed-source APIs by substituting our hardware-grounded latencies with API-measured latencies or per-token pricing.



\section*{Ethical Considerations}
\rw{} is built by aggregating 12 publicly released research benchmarks; we redistribute only data permitted under each source's license and preserve original attribution. By design, the benchmark encourages directing simple queries to small, efficient models rather than defaulting to expensive frontier systems. Widespread adoption of latency-aware routing could meaningfully reduce the energy and compute footprint of large-scale conversational AI, benefiting both service providers and the environment.

At the same time, our findings show that routers trained under partial observability can collapse into degenerate policies that consistently favor a single model, potentially degrading response quality for specific query types or user populations without operators noticing. The Voyager setting we propose is designed to surface exactly these failure modes so that future routing algorithms can be stress-tested before deployment.

\bibliography{custom}

\appendix

\section{Dataset Selection Principles}
\label{app:dataset_principles}

The construction of \rw is guided by the following design principles to ensure a comprehensive and fair assessment of routing capabilities:

\begin{itemize}
    \item \textbf{Capability Diversity:} We include benchmarks spanning six major task categories. Mathematical reasoning, code generation, structured query generation, commonsense and multi-hop reasoning, factual verification and tool use, and open-ended dialogue. This diversity ensures that no single model dominates across all tasks, making routing decisions non-trivial.

    \item \textbf{Discriminative Power:} Each benchmark is selected to induce heterogeneous performance across the model pool. In particular, smaller or specialized models (e.g., \texttt{deepseek-math} on GSM8K or \texttt{codegemma} on APPS) may outperform larger general-purpose models on domain-specific tasks. Such performance divergence is essential for revealing the benefits of intelligent routing over static model selection.

    \item \textbf{Latency Awareness:} Beyond answer correctness, we incorporate real-world serving latency measurements, including time-to-first-token and end-to-end response time. This enables evaluation of quality--latency trade-offs that are critical in practical deployment settings.

    \item \textbf{Reproducibility:} We prioritize benchmarks with well-established evaluation protocols and automatic metrics, avoiding subjective human judgments to ensure consistent and reproducible comparisons across routing methods.

    \item \textbf{Balanced Task Distribution:} Unlike prior benchmarks that are often skewed toward specific capabilities, RouterWorld enforces a balanced distribution across task categories. This design mitigates evaluation bias introduced by over-represented task types and enables fair assessment of routing performance across diverse capabilities.

    \item \textbf{Training--Testing Separation:} Our study evaluates both router models and routing algorithms, so we prioritize benchmarks that provide train and test splits. This separation supports principled learning-based routing and prevents information leakage during evaluation.
\end{itemize}

\section{Additional Figures}
  \label{app:additional-figures}

  \begin{figure*}[h]
  \centering
  \includegraphics[width=1.0\textwidth]{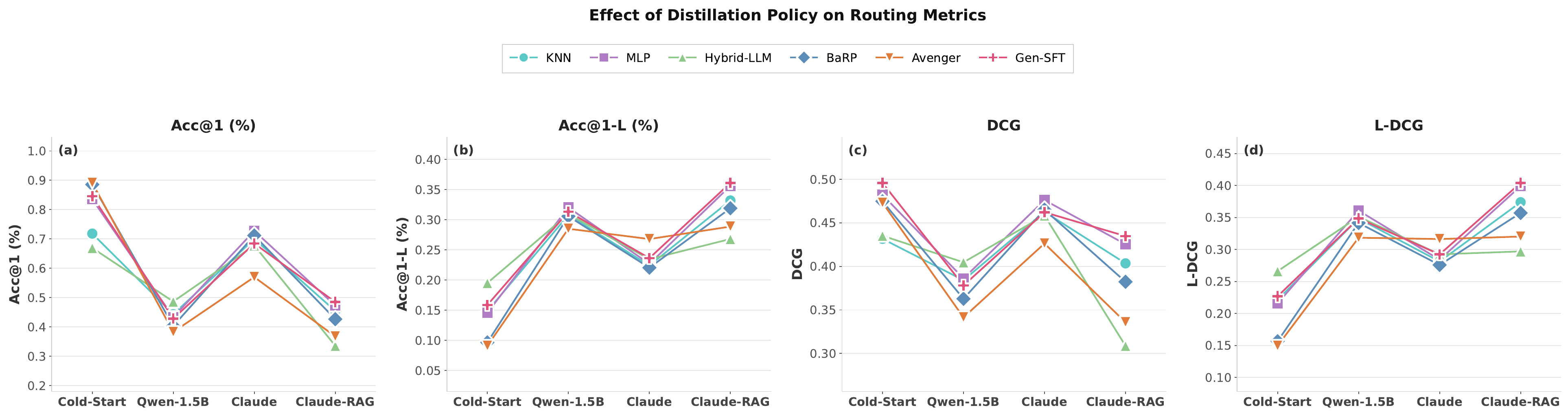}
  \caption{Routing method performance across distillation policies in the partially-observable setting. All methods exhibit a high correlation with the distillation policy trend}
  \label{fig:warmup-wave}
  \end{figure*}

  \begin{figure}[H]
  \centering
  \includegraphics[width=\linewidth]{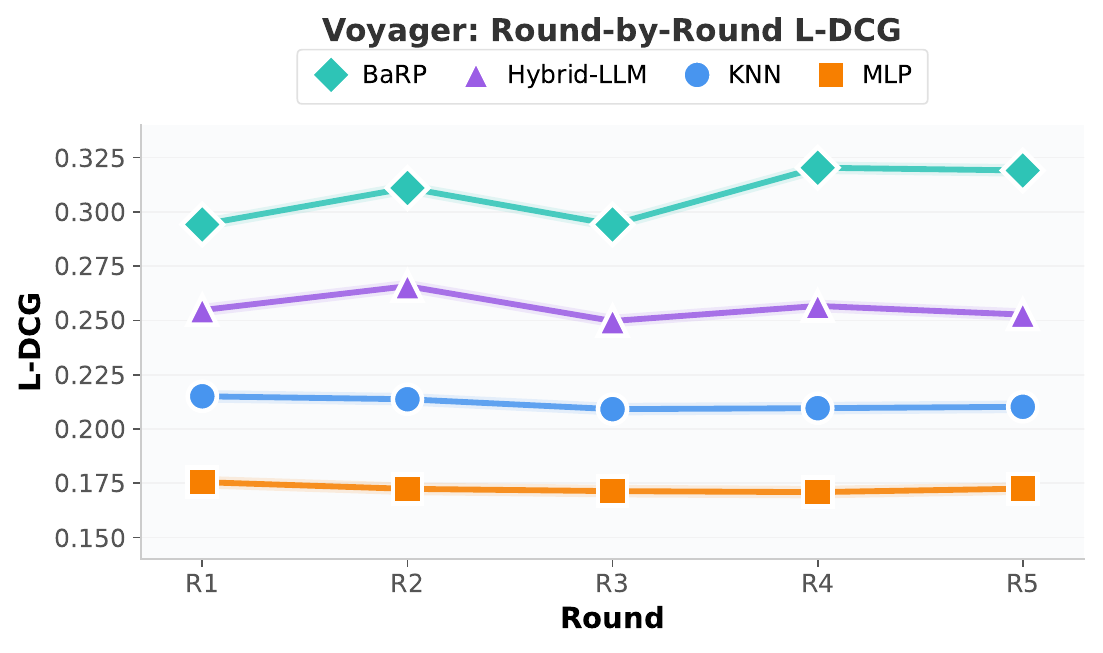}
  \caption{Voyager round-by-round L-DCG.}
  \label{fig:voyager}
  \end{figure}

\section{Benchmark Dataset Details}
\label{app:dataset_details}

The \rw benchmark comprises the following datasets, selected to represent a diverse range of linguistic and symbolic reasoning tasks:

\paragraph{HellaSwag~\citep{hellaswag}.} A sentence completion benchmark that challenges models to select the most plausible continuation of a given context. It employs adversarial filtering to construct distractors that are deceptively plausible to language models while remaining clearly incorrect to humans, effectively testing commonsense reasoning.

\paragraph{GSM8K~\citep{gsm8k}.} A dataset of 8.5K linguistically diverse grade-school math word problems. Each problem typically requires 2--8 steps of elementary arithmetic, testing a model's ability to perform reliable sequential calculations.

\paragraph{ARC-Challenge~\citep{arc_challenge}.} The partition of the AI2 Reasoning Challenge containing grade-school science questions that cannot be solved via simple retrieval or pattern matching. These items require deeper reasoning and scientific understanding.

\paragraph{APPS~\citep{apps}.} A code generation benchmark comprising 10,000 programming problems collected from open-access platforms like Codeforces. The dataset spans three difficulty levels (introductory, interview, and competition) and requires generating executable Python programs from natural language specifications.

\paragraph{BIRD-SQL~\citep{birdsql}.} A large-scale, cross-domain text-to-SQL benchmark featuring real-world databases with noisy and incomplete values. It emphasizes schema understanding and database content reasoning over complex relational structures.

\paragraph{FEVER~\citep{fever}.} A fact verification dataset containing 185K claims derived from Wikipedia. Models must classify claims as \textit{Supported}, \textit{Refuted}, or \textit{NotEnoughInfo} based on retrieved evidence, testing evidence-grounded factual reasoning.

\paragraph{Gorilla APIBench~\citep{gorilla}.} An evaluation of LLM-based API call generation across major hubs (TorchHub, TensorHub, HuggingFace). It tests the ability to produce accurate, executable API invocations from natural language instructions.

\paragraph{HotpotQA~\citep{hotpotqa}.} A multi-hop question answering benchmark requiring reasoning over multiple supporting documents. Questions necessitate combining information from two or more Wikipedia passages, testing compositional reasoning and information synthesis.

\paragraph{PersonaChat~\citep{personachat}.} A chit-chat dialogue dataset where each speaker is assigned a specific persona defined by profile sentences. Models must generate coherent, engaging responses consistent with the assigned persona.

\paragraph{TabMWP~\citep{tabmwp}.} A collection of 38K tabular math word problems combining natural language questions with structured tables. Solutions require extracting relevant numerical information and performing multi-step quantitative reasoning over the tabular data.

\paragraph{Hendrycks-MATH~\citep{hendrycks}.} A dataset of 12,500 competition-level mathematics problems across seven domains, including algebra, geometry, and number theory. These problems demand formal, multi-step mathematical reasoning beyond elementary arithmetic.

\paragraph{MCPToolBench++~\citep{mcptoolbench++}.} Evaluates tool-use capabilities under the Model Context Protocol (MCP). It tests the model's ability to correctly invoke tools via structured function calls and accurately extract parameters from natural language instructions.

\section{Model Specifications and Latency}\label{app:model_details}

\subsection{Latency Profiling Protocol}\label{app:latency-profiling}

To obtain per-model latency estimates for the routing objective, we profile the inference latency of all candidate LLMs using NVIDIA GenAI-Perf, a standardized benchmarking tool built on top of Triton Perf Analyzer.

\paragraph{Setup.}
Each model is served independently via vLLM~\cite{vllm} inside a Docker container with NVIDIA GPU support, using tensor parallelism across 4 GPUs.
The benchmarking client runs inside a separate NVIDIA Triton SDK container (\texttt{tritonserver:24.06-py3-sdk}) on the same host, communicating with the vLLM server over the OpenAI-compatible chat endpoint.
All models are evaluated under identical conditions: 1{,}000 synthetic prompts with a fixed input length of 2{,}000 tokens and output length of 1{,}000 tokens, at a concurrency level of 10, with streaming enabled.

\paragraph{Metrics.}
We extract two key metrics for each model:
\begin{itemize}
\item \textbf{Time to First Token (TTFT)}: the average time from request submission to the first generated token, reflecting model loading and prefill latency.
\item \textbf{Inter-Token Latency (ITL)}: the average time between consecutive output tokens during streaming decoding.
\end{itemize}

\paragraph{Total latency estimation.}
For \rw-Core, where each instance is a single-turn query with $N$ output tokens, we estimate the end-to-end generation latency as:
\begin{equation}
\small
\text{Latency}_{\text{core}}(m, N) = \text{TTFT}_m + (N - 1) \times \text{ITL}_m
\label{eq:latency-core}
\end{equation}
where $m$ denotes the model.

For \rw-Agentic, tasks involve multi-step episodes where the agent interacts with an environment over $K$ turns, each producing $N_k$ tokens. The total episode latency is:
\begin{equation}
\small
\begin{split}
\text{Latency}_{\text{agent}}(m) &= \sum_{k=1}^{K} \bigl[\text{TTFT}_m + (N_k - 1) \times \text{ITL}_m\bigr] \\
& + \sum_{k=1}^{K} \text{Env}_k
\end{split}
\end{equation}
where $\text{Env}_k$ denotes the environment response time at step $k$ (e.g., browser rendering, API calls). In practice, we record the wall-clock episode latency directly from the agentic harness, which captures both model inference and environment interaction time.

These formulas are used to compute per-instance latency for each model in the routing dataset, enabling the construction of latency-aware ground-truth labels.

\subsection{Unified Latency Profile}

Table~\ref{tab:latency-profile} presents the latency profile of all candidate models across both \rw-Core and \rw-Agentic, measured with GenAI-Perf under identical conditions. Models are sorted by total estimated latency (Eq.~\ref{eq:latency-core} with $N{=}1{,}000$). The \textbf{Pool} column indicates whether the model is used in Core, Agentic, or both benchmarks.

\begin{table*}[t]
\centering
\setlength{\tabcolsep}{4pt}
\begin{tabular}{llrrrr}
\toprule
\textbf{Model} & \textbf{Pool} & \textbf{Params} & \textbf{TTFT (ms)} & \textbf{ITL (ms)} & \textbf{Total (s)} \\
\midrule
Gemma-2B              & Core    & 2B   &      12.64 &  3.96 &  3.97 \\
CodeGemma-7B          & Core    & 7B   &      14.22 &  5.20 &  5.21 \\
Ministral-14B         & Agentic & 14B  &      18.26 &  3.50 &  3.51 \\
Mistral-7B            & Core    & 7B   &     100.21 &  4.13 &  4.23 \\
DS-Math-7B            & Core    & 7B   &     117.16 &  4.36 &  4.48 \\
Devstral-123B         & Agentic & 123B &     493.30 & 14.56 & 15.04 \\
Qwen3-Coder-30B       & Agentic & 30B  &     532.29 &  7.20 &  7.73 \\
GPT-OSS-20B           & Both    & 20B  &  1{,}137.45 &  4.06 &  5.19 \\
GPT-OSS-120B          & Agentic & 120B &  1{,}212.55 &  5.97 &  7.18 \\
Qwen3-235B            & Both    & 235B &  2{,}258.10 & 18.76 & 21.01 \\
Nemotron-9B           & Core    & 9B   &  5{,}733.59 &  6.29 & 12.03 \\
Qwen3-Next-80B        & Agentic & 80B  & 20{,}612.40 &  8.38 & 28.99 \\
\bottomrule
\end{tabular}
\caption{Unified latency profile of all candidate models, measured with GenAI-Perf (2{,}000 input tokens, 1{,}000 output tokens, concurrency 10, streaming). Models sorted by total latency.}
\label{tab:latency-profile}
\end{table*}

\subsection{Model Descriptions}

\paragraph{Core-Only Models.}
\textbf{Gemma-2B} serves as the efficiency anchor with a median latency of just 13ms TTFT. \textbf{CodeGemma-7B} and \textbf{DS-Math-7B} are domain-specialized models that test the router's ability to identify task intent, offering low latency (${\sim}$100ms) while outperforming generalists on code and math tasks. \textbf{Mistral-7B} provides a robust generalist baseline for the 7B class.

\paragraph{Agentic-Only Models.}
\textbf{Ministral-14B} is the fastest agentic model (18ms TTFT, 3.5ms ITL). \textbf{Qwen3-Coder-30B} represents a mid-tier option balancing latency and accuracy. \textbf{Devstral-123B} and \textbf{GPT-OSS-120B} are large-scale models for complex agentic tasks. \textbf{Qwen3-Next-80B} exhibits the highest TTFT (20.6s) due to cold-start overhead.

\paragraph{Shared Models.}
\textbf{GPT-OSS-20B} bridges both pools, achieving 67.9\% accuracy on Core with moderate latency (${\sim}$5.2s). \textbf{Qwen3-235B} is the oracle model of both pools with the highest accuracy but significant latency (${\sim}$21s). \textbf{Nemotron-9B} achieves high accuracy (71.8\% on Core) but exhibits disproportionately high TTFT (5.7s), presenting a challenging routing trap: high accuracy at high cost.

\paragraph{Infrastructure.}
All profiling experiments are conducted on a single node equipped with NVIDIA GPUs. Models requiring newer architecture support (Ministral-14B, Qwen3-Next-80B, Devstral-123B) use vLLM v0.12.0, while the remaining models use vLLM v0.9.2, all with tensor parallelism of 4.

\section{Detailed Dataset Statistics \& Analysis}\label{app:dataset_stats}

\subsection{\rw-Core Statistics}\label{app:dataset_stats_core}

In this section, we analyze the specific difficulty distributions and routing opportunities within \rw-Core and \rw-Agentic.

\begin{figure*}[h]
\centering
\includegraphics[width=0.95\textwidth]{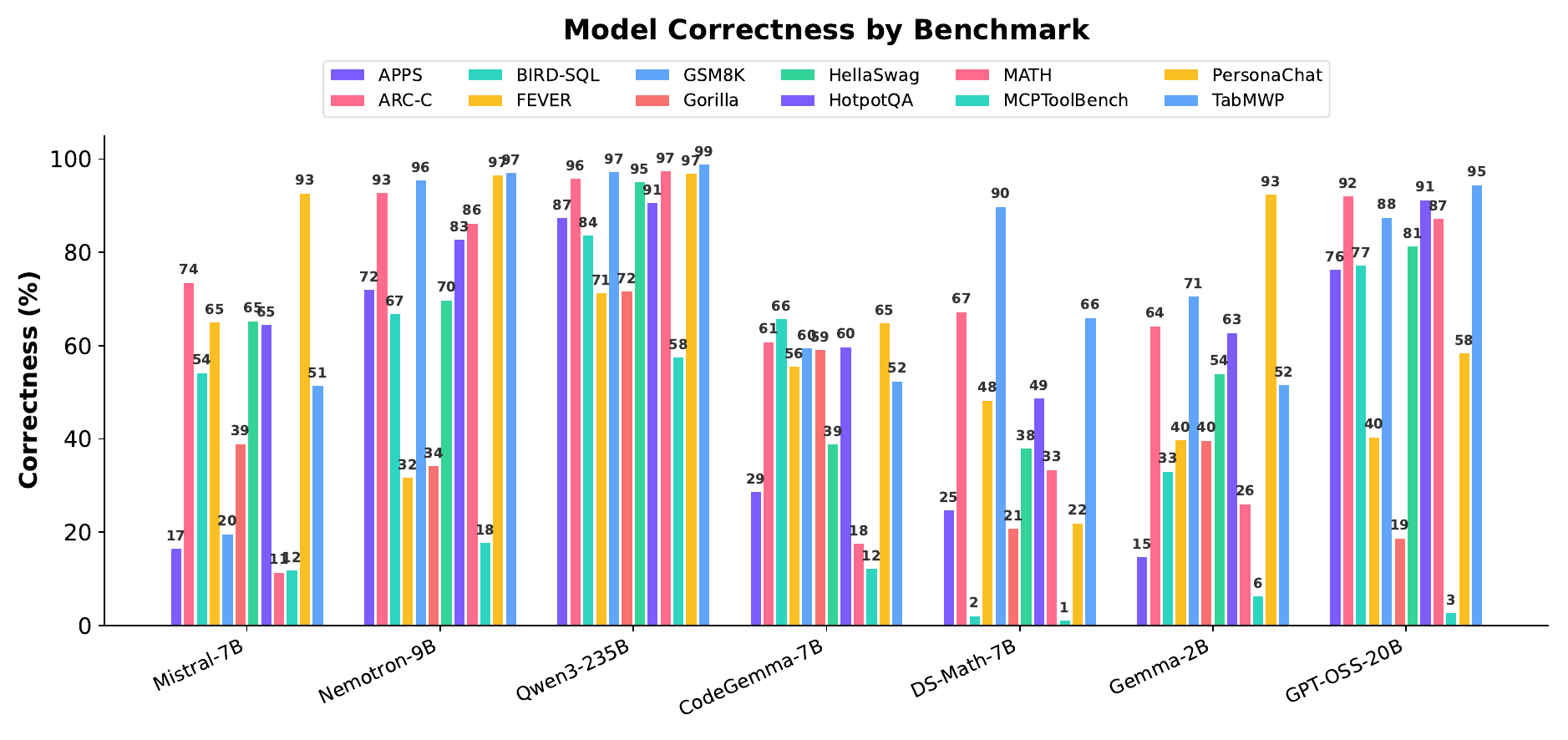}
\caption{Per-benchmark model correctness for \rw-Core. Models exhibit
distinct specialization patterns across task types, with smaller models
often outperforming generalists in specific domains.}
\label{fig:core_correctness}
\end{figure*}

\paragraph{Composition and Specialization.}
The \rw-Core test set ($N=7{,}967$) provides balanced coverage across
Reasoning (Logic 17.3\%, Math 18.5\%, Data Analysis 18.9\%) and
Interaction (QA 16.3\%, Code 12.0\%, Tool Use 10.6\%, Chit-Chat 6.4\%).
As shown in Figure~\ref{fig:core_correctness}, no single model dominates
all categories. For example, \texttt{DS-Math-7B} leads sub-20B models on
Math tasks but underperforms on Logic, while \texttt{CodeGemma-7B} shows
distinct strength in code generation. This confirms that effective routing
must account for task-specific competence rather than relying solely on
model scale.

\begin{figure*}[h]
\centering
\begin{minipage}[t]{0.48\textwidth}
\centering
\includegraphics[width=\textwidth]{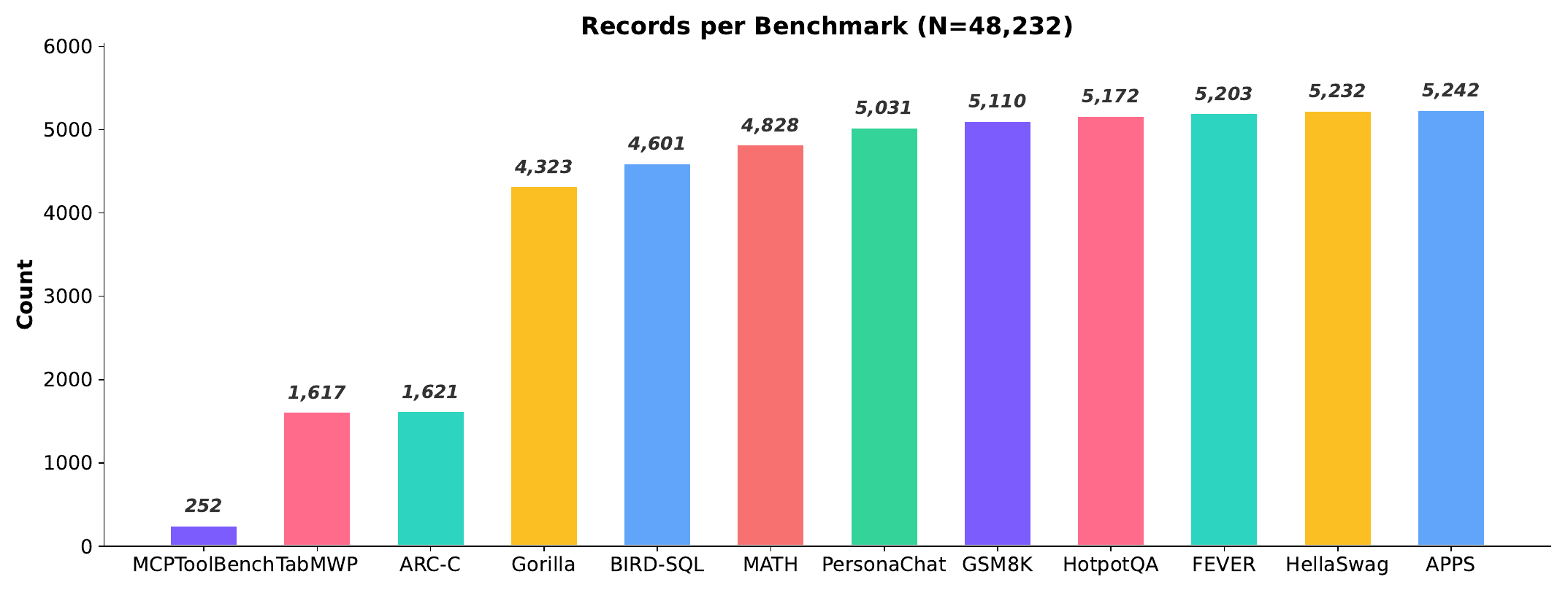}
\caption{Records per benchmark in \rw-Core. The dataset spans 12 benchmarks
with broad coverage across reasoning, coding, and interaction tasks.}
\label{fig:core_records}
\end{minipage}
\hfill
\begin{minipage}[t]{0.48\textwidth}
\centering
\includegraphics[width=\textwidth]{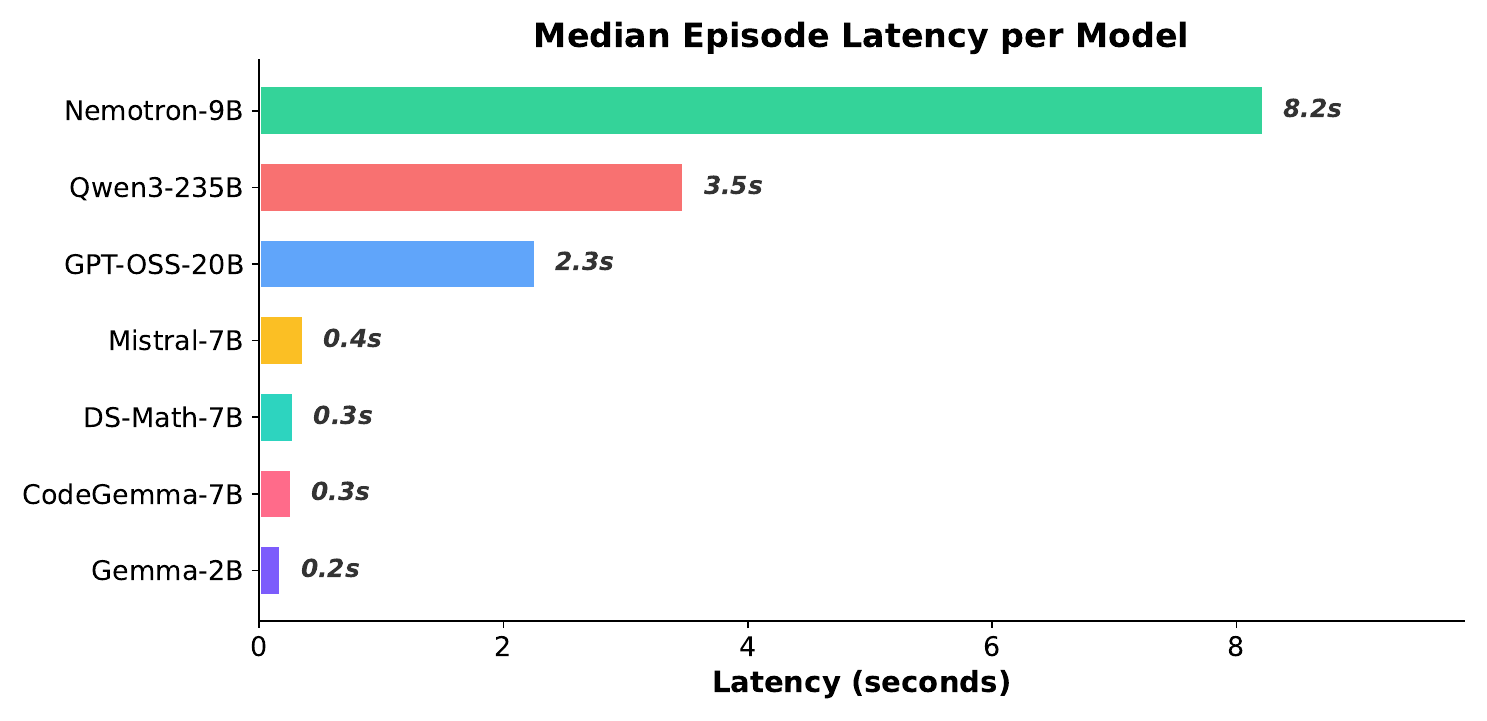}
\caption{Median episode latency per model in \rw-Core. Smaller models
offer substantially lower latency, motivating intelligent routing.}
\label{fig:core_latency}
\end{minipage}
\end{figure*}

\begin{figure*}[h]
\centering
\includegraphics[width=\textwidth]{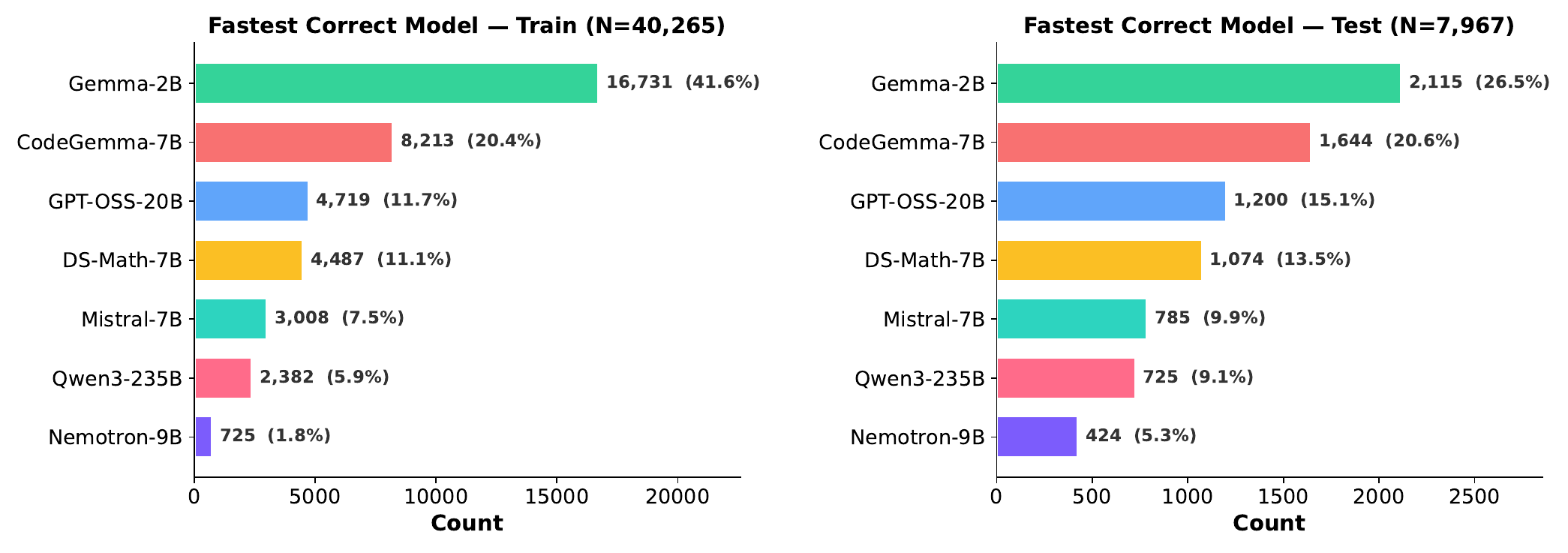}
\caption{Fastest correct model distribution for \rw-Core (train and test).
Small models are the optimal choice for a large share of queries.}
\label{fig:core_fastest}
\end{figure*}

\paragraph{Routing Opportunity.}
Query difficulty is broadly distributed (Figure~\ref{fig:core_records}),
creating significant opportunity for adaptive routing.
Oracle routing selects the \textit{fastest correct} model per query,
demonstrating that frontier-scale models are often computationally
wasteful for standard tasks (Figure~\ref{fig:core_fastest}).

\subsection{\rw-Agentic Statistics}\label{app:dataset_stats_agentic}

\paragraph{Composition.}
The \rw-Agentic benchmark ($N_{\text{train}}=11{,}292$,
$N_{\text{test}}=1{,}448$) covers three agentic task families:
Web Navigation (44.8\%), Online Shopping (38.2\%), and Household (17.0\%),
drawn from WorkArena, WebShop, and ALFWorld respectively.
These tasks require multi-step reasoning, tool use, and environment
interaction, presenting fundamentally different routing challenges
compared to single-turn Core tasks.

\begin{figure*}[!t]
\centering
\includegraphics[width=0.95\linewidth]{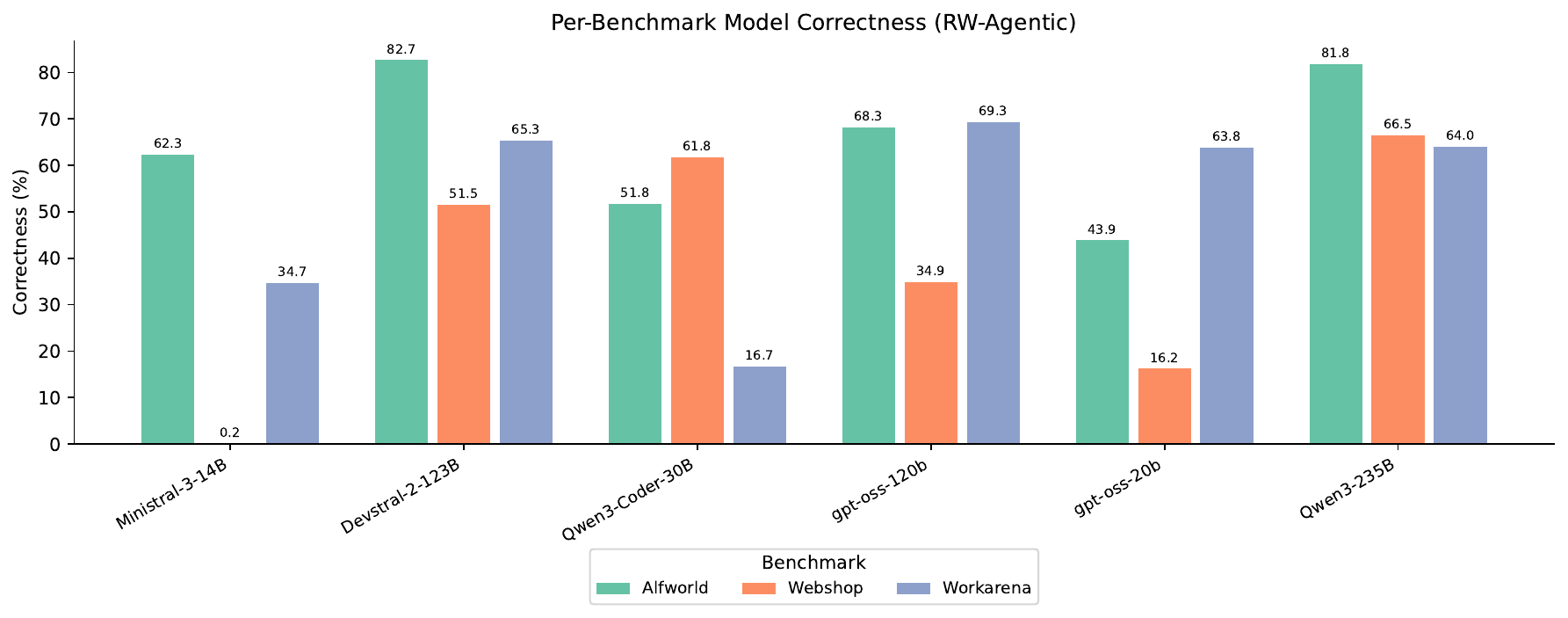}
\caption{Per-benchmark model correctness for \rw-Agentic. Performance
varies substantially across the three agentic task families.}
\label{fig:agent_correctness}
\end{figure*}

\paragraph{Model Correctness by Benchmark.}
As shown in Fig.~\ref{fig:agent_correctness}, correctness varies substantially across benchmarks. Larger models achieve higher scores on complex web navigation tasks, but smaller specialized models can be competitive on structured household tasks. This heterogeneity
across benchmarks reinforces the value of task-aware routing.

\begin{figure}[t]
\centering
\includegraphics[width=\linewidth]{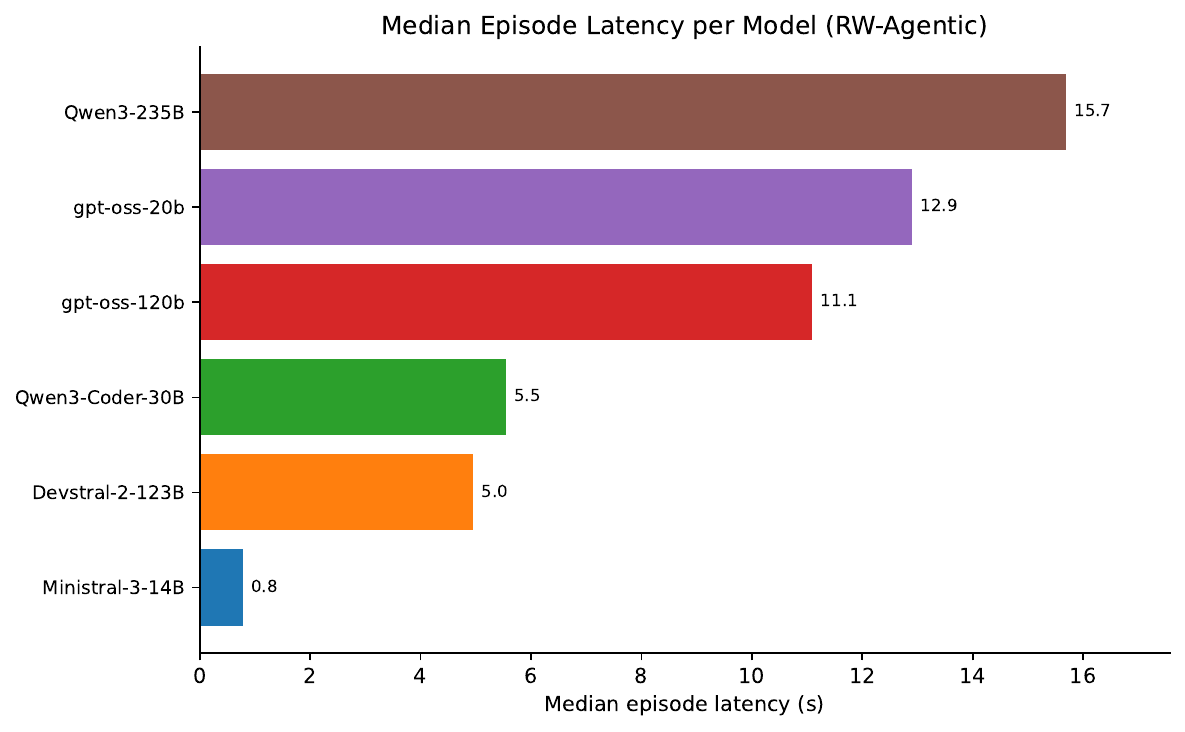}
\caption{Median episode latency per model in \rw-Agentic. Latency
ranges from 0.8s to 15.7s across models.}
\label{fig:agent_latency}
\end{figure}

\begin{figure*}[h]
\centering
\includegraphics[width=0.95\textwidth]{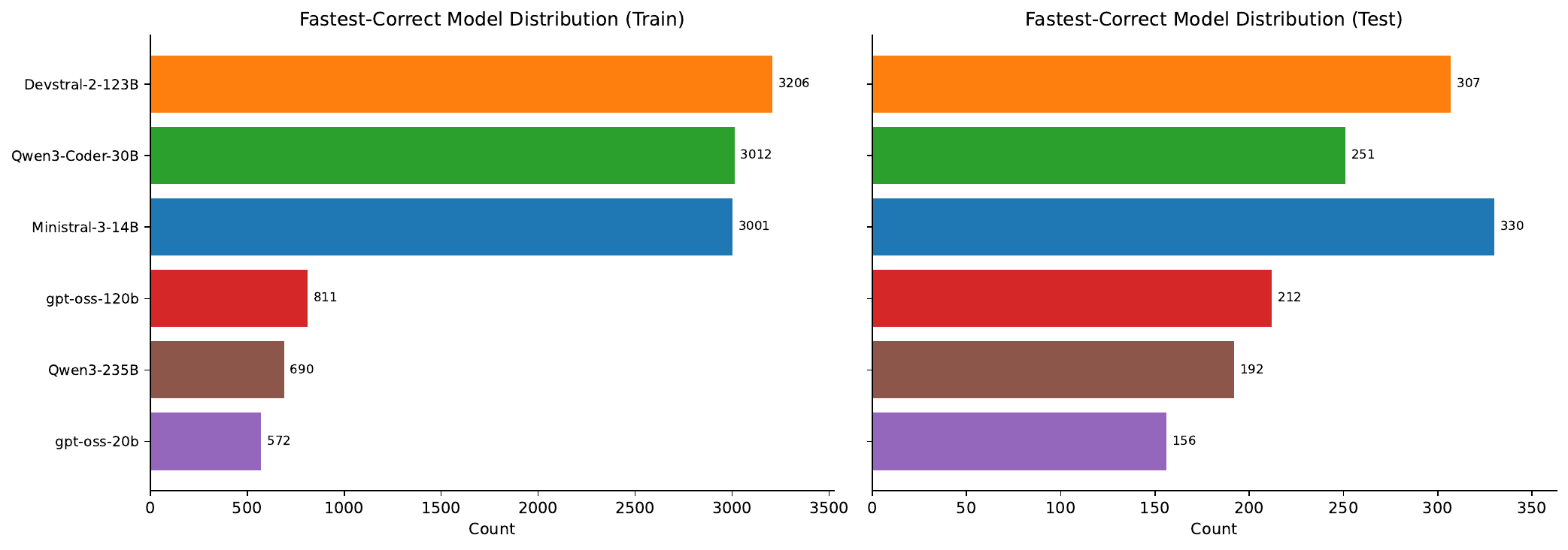}
\caption{Fastest correct model distribution for \rw-Agentic (train and test).}
\label{fig:agent_fastest}
\end{figure*}

\paragraph{Latency and Routing Opportunity.}
Agentic tasks exhibit significantly higher latency than Core tasks (median 11.8s vs.\ 4.3s), with large variance across models (Figure~\ref{fig:agent_latency}). This amplifies the potential savings from intelligent routing. Figure~\ref{fig:agent_fastest} shows the
fastest correct model distribution, revealing that even in the agentic
setting, smaller models frequently provide the fastest correct response.

\begin{figure}[t]
\centering
\includegraphics[width=0.8\linewidth]{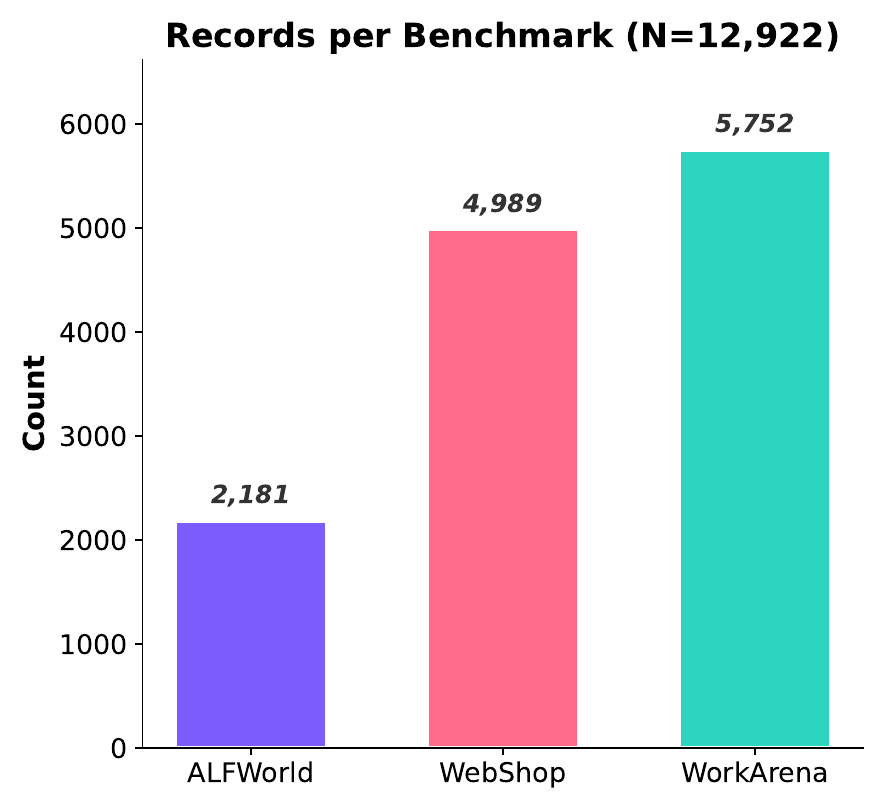}
\caption{Records per benchmark in \rw-Agentic across three agentic
task families.}
\label{fig:agent_records}
\end{figure}

\section{Router Implementation}
\label{app:algorithm}

\subsection{Hyperparameters}
The regression head is implemented as a three-layer Multi-Layer Perceptron (MLP) with hidden dimensions $[512, 256, 128]$. Each linear layer is followed by a ReLU activation, Batch Normalization, and Dropout ($p=0.3$). The final projection maps the hidden state to $K$ logits. We train using the AdamW optimizer with a learning rate of $2\times 10^{-5}$ for the encoder and $1\times 10^{-4}$ for the head, employing a cosine annealing schedule with a 5\% warmup period.

\section{Baseline Implementation Details}\label{app:baselines}

We implement all baselines using PyTorch and train on $8\times$ NVIDIA H200 GPUs.
Below we detail the architecture and hyperparameters for each method under both the
\emph{fully observable} and \emph{partially observable} settings.
In the fully observable setting, the router has access to the correctness labels of
\emph{all} candidate models for each training query (\texttt{correct\_list}).
In the partially observable setting, the router only observes the outcome of a
\emph{single} selected model per query (\texttt{selected} field), and must learn to
route without knowledge of unobserved model outcomes.
Unless otherwise noted, all methods share the same architecture across settings;
we highlight where hyperparameters or training procedures differ.

\subsection{Representation-Based Routers}

Representation-based routers learn a query representation and make routing decisions
based on similarity, classification, or learned utility in embedding space. These
methods differ in whether representations are frozen or fine-tuned, and in how routing
decisions are derived from the learned representations.

\begin{itemize}

\item \textbf{Naive Bayes \& KNN ($k{=}5$):}
Operate on frozen representations from \texttt{Qwen3-Embedding-0.6B}
(mean-pooled, 1024-dimensional), predicting a distribution over the $K$ candidate
models and selecting the argmax.
These methods are non-parametric and require no gradient-based training.
\begin{itemize}
  \item \emph{Fully observable:} Trained on the full training set with labels derived
    from \texttt{correct\_list}.
  \item \emph{Partially observable:} Trained on the filtered subset containing only
    samples where the selected model was correct (\texttt{correct\_only}).
\end{itemize}

\item \textbf{MLP:}
A three-hidden-layer feedforward classifier ($1024 \to 512 \to 256 \to 128 \to K$)
trained on frozen \texttt{Qwen3-Embedding-0.6B} representations using cross-entropy
loss (learning rate $1\times10^{-4}$, dropout $0.2$).
\begin{itemize}
  \item \emph{Fully observable:} Trained for 12 epochs on the full training set.
  \item \emph{Partially observable:} Trained for 20 epochs on the \texttt{correct\_only}
    subset.
\end{itemize}

\item \textbf{Emb-FT:}
End-to-end fine-tuning of \texttt{Qwen3-Embedding-0.6B} with a two-layer
classification head (hidden dimension 512, dropout $0.1$), trained with DDP across
8 GPUs (learning rate $1\times10^{-4}$, per-GPU batch size 32, gradient accumulation
2, effective batch size 512).
\begin{itemize}
  \item \emph{Fully observable:} 10 epochs on the full training set.
  \item \emph{Partially observable:} 10 epochs on the \texttt{correct\_only} subset,
    with labels from the \texttt{selected} field.
\end{itemize}

\item \textbf{ModernBERT}~\cite{modernbert}:
Fine-tunes \texttt{ModernBERT-base} end-to-end with a two-layer classification head
(hidden dimension 768, dropout $0.1$) using DDP across 8 GPUs.
\begin{itemize}
  \item \emph{Fully observable:} Learning rate $1\times10^{-3}$, per-GPU batch size 32,
    gradient accumulation 2, effective batch size 512, 10 epochs.
  \item \emph{Partially observable:} Learning rate $1\times10^{-3}$, per-GPU batch size
    128, gradient accumulation 2, effective batch size 2{,}048, 10 epochs on the
    \texttt{correct\_only} subset.
\end{itemize}

\item \textbf{Hybrid-LLM}~\cite{hybridllm}:
Fine-tunes \texttt{DeBERTa-v3-large} with an $N$-class cross-entropy objective
(learning rate $5\times10^{-5}$, dropout $0.1$) using DDP across 8 GPUs.
\begin{itemize}
  \item \emph{Fully observable:} Per-GPU batch size 32, gradient accumulation 2,
    effective batch size 512, 20 epochs. Uses \texttt{correct\_list[0]} (fastest
    correct model) as the ground-truth label.
  \item \emph{Partially observable:} Per-GPU batch size 64, gradient accumulation 2,
    effective batch size 1{,}024, 15 epochs on the \texttt{correct\_only} subset.
    Uses the \texttt{selected} model as the ground-truth label.
\end{itemize}

\item \textbf{BaRP}~\cite{barp}:
A preference-conditioned routing policy trained with REINFORCE on top of frozen
\texttt{Qwen3-Embedding-0.6B} representations (learning rate $1\times10^{-4}$,
entropy coefficient $\beta{=}0.05$, dropout $0.2$). At inference, users specify a
preference vector $(w_q, w_c)$ to trade off quality and cost; the policy outputs
$\pi(a \mid x, w)$ over candidate models. Trained with DDP across 8 GPUs
(per-GPU batch size 32, effective batch size 256).
\begin{itemize}
  \item \emph{Fully observable:} Uses expected reward over all models from
    \texttt{correct\_list}, up to 100 epochs with early stopping (patience 10).
  \item \emph{Partially observable:} Learns only from the outcome of the
    \texttt{selected} model per sample, up to 100 epochs with early stopping
    (patience 10).
\end{itemize}

\item \textbf{GraphRouter}~\cite{graphrouter}:
A graph neural network modeling query--model interactions. GNN layers
(hidden dimension 8, edge mask rate 0.5) are trained on top of frozen
\texttt{Qwen3-Embedding-0.6B} embeddings with a learning rate of $3\times10^{-4}$
and weight decay $1\times10^{-4}$.
\begin{itemize}
  \item \emph{Fully observable:} Batch size 128, 100 epochs, single GPU. Labels
    derived from \texttt{correct\_list[0]} (fastest correct model).
  \item \emph{Partially observable:} Batch size 128, 1{,}000 epochs, single GPU.
    Only the selected model receives gradient (mask${}=1$); unobserved models are
    masked out.
\end{itemize}

\item \textbf{Avenger}~\cite{avenger}:
A \emph{training-free}, cluster-based ranking router. During offline calibration,
training queries are embedded with frozen \texttt{Qwen3-Embedding-0.6B} and clustered
via K-means ($k{=}64$). Per-cluster model rankings are computed from the training
labels. At inference, the query is assigned to the top-$11$ nearest clusters
(softmax temperature $\beta{=}3.0$), and rankings are aggregated to select the
best model.
\begin{itemize}
  \item \emph{Fully observable:} Rankings computed using L-DCG (latency-aware
    discounted cumulative gain) over all models in \texttt{correct\_list}.
  \item \emph{Partially observable:} Rankings computed using \texttt{router\_accuracy}
    over the \texttt{selected} field only, as latency-aware metrics require
    observing all models.
\end{itemize}

\end{itemize}

\subsection{Generative Routers}

Generative routers use autoregressive language models to directly select a target model
via prompting or supervised fine-tuning on routing data.

\begin{itemize}

\item \textbf{Zero-Shot Prompting:}
We serve \texttt{Qwen2.5-0.5B-Instruct} via vLLM with 128 parallel workers. The model
receives a system prompt describing the available model set and outputs the selected
model name directly. This baseline is identical across both settings as it requires
no training data.

\item \textbf{Gen-SFT:}
Supervised fine-tuning of \texttt{Qwen2.5-1.5B-Instruct} on routing annotations using
standard cross-entropy loss (learning rate $2\times10^{-5}$, micro batch size 4,
effective batch size 128, max sequence length 1{,}024, weight decay $0.01$, 5 epochs,
8 GPUs with FSDP).
\begin{itemize}
  \item \emph{Fully observable:} Training targets are the fastest correct model from
    \texttt{correct\_list}.
  \item \emph{Partially observable:} Training targets are the \texttt{selected} model,
    filtered to correct-only samples.
\end{itemize}

\end{itemize}

\section{LLM Judge Validation Details}
\label{app:judge-validation}

We validate the automated LLM judge used throughout \rw against independent
human annotations.
Six experts blind-label 150 stratified (query, model, response) triples
sampled across 12~datasets and 7~models without seeing the judge verdict.
We report inter-annotator agreement (IAA) to establish the task-difficulty
ceiling, and LLM--human agreement (LHA) to measure judge alignment with
expert majority-vote consensus.

\begin{figure}[t]
\centering
\includegraphics[width=\linewidth]{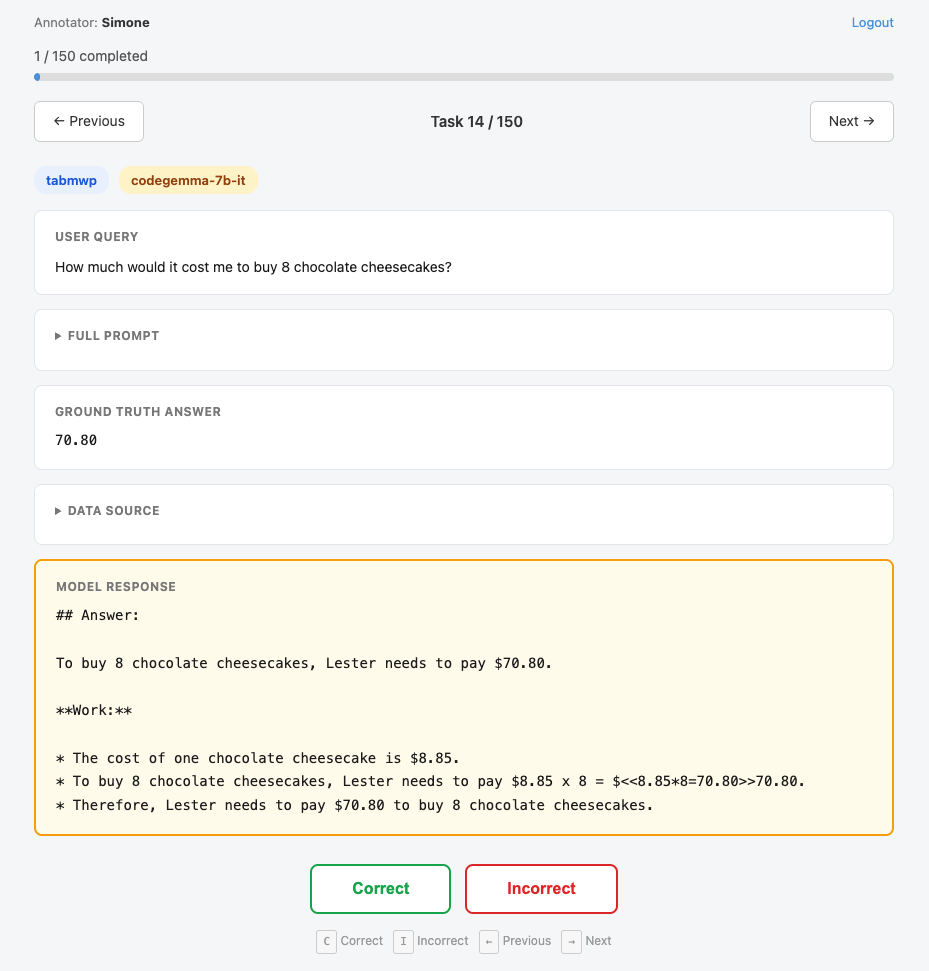}
\caption{Screenshot of the web-based annotation tool developed for the
  LLM judge validation study.
  The interface displays the source dataset and model (colored tags),
  the user query, the full prompt, the ground-truth answer, and the
  model response.
  Annotators label each triple as \emph{correct} or \emph{incorrect}
  by comparing the model response against the ground truth.
  The tool and annotation guidelines will be released as open source.}
\label{fig:annotation-tool}
\end{figure}

\subsection{Annotation Protocol}
\label{app:annotation-guidelines}

We developed a web-based annotation tool (Figure~\ref{fig:annotation-tool})
that presents each annotator with:
(i)~the user query,
(ii)~the full prompt including instructions and response choices,
(iii)~the ground-truth answer from the dataset, and
(iv)~the model response.
The source dataset and responding model are displayed as colored tags.
Annotators follow the guidelines below.

\paragraph{Core Rule.}
For each triple, the annotator compares the model response against the
ground-truth answer and labels it \emph{correct} or \emph{incorrect}.

\paragraph{Key Guidelines.}
\begin{enumerate}
  \item \textbf{Focus on the final answer only.}
        Formatting differences (e.g., \texttt{<a>C</a>} vs.\ \texttt{C.})
        and intermediate reasoning traces are ignored;
        only the substantive answer matters.
  \item \textbf{Compare against ground truth.}
        Annotators judge whether the model's answer matches the provided
        ground-truth label, without independently verifying the ground truth
        itself.
  \item \textbf{Skip ambiguous cases.}
        When a confident judgment is not possible---e.g., long code-generation
        tasks requiring execution, or subjective dialogue evaluations
        (persona-chat)---annotators skip the entry to preserve annotation
        integrity.
\end{enumerate}

\subsection{Agreement Scores}

Table~\ref{tab:app-iaa} reports multi-rater agreement among the six annotators.
The high Fleiss' $\kappa{=}0.84$ reflects the largely objective nature of the
benchmarks: for datasets with unambiguous answers
(e.g., multiple-choice, numeric), annotators almost always concur.
Disagreements arise primarily on tasks requiring non-trivial answer extraction
(e.g., \textsc{bird-sql}, \textsc{gorilla\_func}).

\begin{table}[t]
\centering
\begin{tabular}{lc}
\toprule
Metric & Value \\
\midrule
Fleiss' $\kappa$ & 0.835 \\
Krippendorff's $\alpha$ & 0.867 \\
Avg.\ pairwise $\kappa$ & 0.892 \\
Avg.\ pairwise agreement & 0.950 \\
\bottomrule
\end{tabular}
\caption{Inter-annotator agreement among 6 experts on 150 blind annotations.}
\label{tab:app-iaa}
\end{table}

Table~\ref{tab:app-agreement} compares the majority-vote consensus with the
LLM judge.
The judge--consensus $\kappa{=}0.84$ matches the inter-expert Fleiss'
$\kappa{=}0.84$, indicating the judge operates at human-level reliability,
consistent with the findings of~\cite{zheng2023judging} who report $>$80\%
agreement between GPT-4 and human evaluators.

\begin{table}[t]
\centering
\resizebox{\linewidth}{!}{%
\begin{tabular}{lcccc}
\toprule
Comparison & Agreement & $\kappa$ & Precision / Recall & F1 \\
\midrule
Consensus vs.\ LLM Judge & 0.920 & 0.835 & 0.98\,/\,0.89 & 0.932 \\
\bottomrule
\end{tabular}%
}
\caption{Majority-vote consensus (6~annotators) vs.\ LLM judge ($n{=}150$).}
\label{tab:app-agreement}
\end{table}

\subsection{Individual Annotator vs.\ LLM Judge}

Table~\ref{tab:app-individual} reports each annotator's agreement with the
judge, showing that the judge aligns comparably with every expert
($\kappa$ range: 0.73--0.93).

\begin{table}[t]
\centering
\begin{tabular}{lccc}
\toprule
Annotator & $n$ & Agreement & $\kappa$ \\
\midrule
Annotator-S    & 87  & 94.3\% & 0.880 \\
Annotator-A     & 86  & 94.2\% & 0.871 \\
Annotator-D     & 88  & 96.6\% & 0.929 \\
Annotator-H    & 100 & 89.0\% & 0.772 \\
Annotator-P   & 100 & 87.0\% & 0.730 \\
Annotator-W   & 97  & 93.8\% & 0.874 \\
\bottomrule
\end{tabular}
\caption{Individual annotator agreement with the LLM judge on the 150-sample set.}
\label{tab:app-individual}
\end{table}

\section{Detailed Results on \rw-Agentic}
\label{app:agentic_results}

\subsection{Agentic Benchmark Details}
\label{app:agentic_details}

\textsc{RW-Agentic} evaluates routing across three interactive benchmarks: ALFWorld (household tasks), WebShop (web navigation), and WorkArena (enterprise workflows), totaling over 76{,}000 model--task interactions.
The candidate pool comprises six models spanning 14B--235B parameters:
Ministral-14B, Qwen3-Coder-30B, Devstral-123B, GPT-OSS-20B, GPT-OSS-120B, and Qwen3-235B.
Each instance records the user query, the model's step-by-step interaction trace, task success, number of steps, and cumulative episode latency.

Table~\ref{tab:agentic_model_pool} summarizes per-model statistics.
Unlike \textsc{RW-Core} where latency is dominated by single-turn inference, agentic latency reflects the full episode duration, a product of per-step inference cost and the number of interaction steps required.

\begin{table}[t]
\centering
\resizebox{\linewidth}{!}{%
\begin{tabular}{lrrrr}
\toprule
\textbf{Model} & \textbf{TTFT (ms)} & \textbf{ITL (ms)} & \textbf{Throughput (tok/s)} & \textbf{Med.\ Episode Lat.\ (s)} \\
\midrule
Ministral-14B       &    18 &  3.5 & 2{,}764 &  0.6 \\
Qwen3-Coder-30B     &   532 &  7.2 & 1{,}126 &  5.5 \\
Devstral-123B       &   493 & 14.6 &   574   &  4.9 \\
GPT-OSS-20B         & 1{,}137 &  4.1 &   987 & 12.7 \\
GPT-OSS-120B        & 1{,}213 &  6.0 & 1{,}169 & 11.1 \\
Qwen3-235B          & 2{,}258 & 18.8 &   370   & 16.4 \\
\bottomrule
\end{tabular}%
}
\caption{Agentic candidate model pool. TTFT = time to first token; ITL = inter-token latency. Episode latency is the median total time across all tasks.}
\label{tab:agentic_model_pool}
\end{table}

\subsection{Performance Compression and Policy Collapse}
\label{app:agentic_compression}

Under the 3-model pool, fully-observable methods cluster tightly with near-zero Acc@1-L variance ($\sigma^2{=}0.81$), compared to $\sigma^2{=}29.5$ in \textsc{RW-Core}.
Expanding to 6 models does not alleviate this---among non-collapsed methods, variance remains low ($\sigma^2{=}4.35$ for Acc@1-L in fully-observable), and the gap to the Oracle upper bound (70.65\% Acc@1-L) remains substantial.
Under partial observability, the compression worsens further, with multiple routers collapsing into degenerate single-model policies (Table~\ref{tab:router_comparison_agentic}).

The root cause lies in the interaction between model planning capability and episode length.
Models with stronger planning capabilities resolve tasks in fewer steps (Figure~\ref{fig:agent_steps_full}): Qwen3-235B requires a median of 18 steps on ALFWorld, while GPT-OSS-20B requires 50.
Despite Qwen3-235B having the highest per-step inference latency (15.7s median episode latency), its superior planning yields lower \emph{total} episode cost than weaker models that compensate with many cheap but inefficient steps (Figure~\ref{fig:agent_latency_full}).
This inverts the conventional latency--accuracy trade-off: the largest model is simultaneously the most accurate \emph{and} the most cost-efficient, leaving routers with no meaningful trade-off to exploit.

Without counterfactual feedback, partially-observable routers greedily exploit this dynamic, collapsing into static ``Route-To-Best'' policies.
In the 3-model setting, collapsed routers default to Qwen3-235B; in the 6-model pool, the pattern persists with routers favoring the dominant model.
This collapse is rational given the reward structure---the strongest planner minimizes both error and accumulated latency---but it renders routing trivial.

Breaking this plateau requires routing algorithms that reason about multi-step execution dynamics rather than treating agentic tasks as single-turn queries.
Capturing non-linear planning--latency interactions demands large-scale rollout environments for exploration-guided training.
By providing over 76{,}000 standardized agentic interactions across 6 models, \textsc{RW-Agentic} establishes the largest-scale environment to date for studying these routing bottlenecks in realistic agent deployments.

\subsection{Planning Capability and Step Efficiency}
\label{app:agentic_planning}

\begin{figure}[t]
    \centering
    \includegraphics[width=\linewidth]{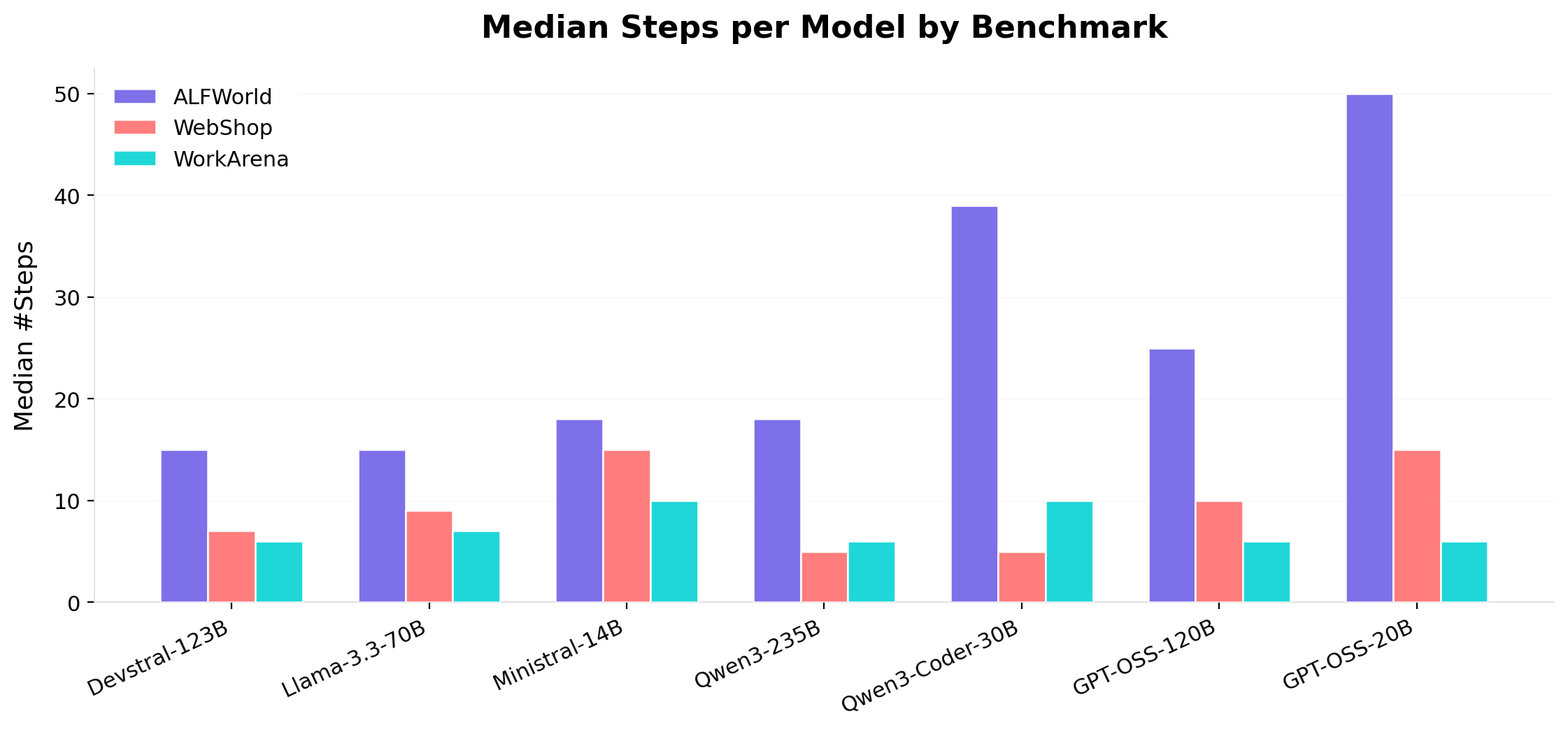}
    \caption{Median interaction steps per model by benchmark. Models with stronger planning capabilities (e.g., Qwen3-235B, Devstral-123B) resolve tasks in fewer steps, particularly on ALFWorld and WebShop.}
    \label{fig:agent_steps_full}
\end{figure}

Figure~\ref{fig:agent_steps_full} shows the median number of interaction steps per model across all three benchmarks.
The relationship between model capability and step efficiency is benchmark-dependent:
on ALFWorld, GPT-OSS-20B requires a median of 50 steps while most other models cluster around 15--18;
on WebShop, Qwen3-235B and Qwen3-Coder-30B are the most step-efficient (median 5 steps);
on WorkArena, all models converge to 5--10 steps.
These patterns confirm that planning capability---not raw inference speed---is the primary driver of episode cost in agentic settings.

\begin{figure}[t]
    \centering
    \includegraphics[width=1\linewidth]{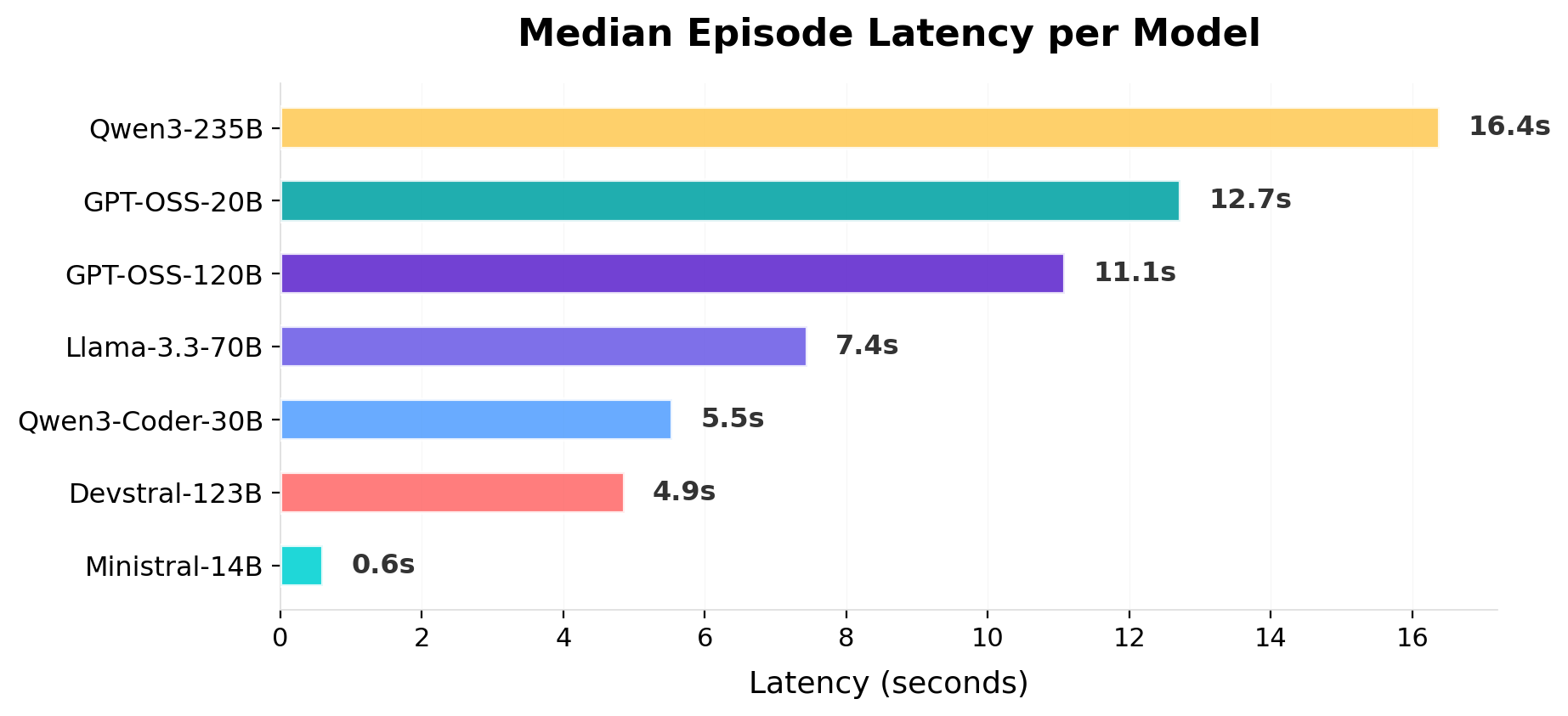}
    \caption{Median episode latency per model. The conventional assumption that larger models are slower does not hold in agentic settings. Planning efficiency can offset higher per-step inference cost.}
    \label{fig:agent_latency_full}
\end{figure}

Figure~\ref{fig:agent_latency_full} shows the resulting median episode latency.
Despite having the highest per-token cost (18.8\,ms ITL), Qwen3-235B achieves competitive total episode latency because its superior planning reduces the number of required steps.
Conversely, GPT-OSS-20B has low per-token cost (4.1\,ms ITL) but incurs the second-highest episode latency due to verbose, multi-step execution.

\subsection{Full Routing Results}
\label{app:agentic_full_results}

\subsubsection{6-Model Pool}

Table~\ref{tab:router_comparison_7way} presents routing results with the full 6-model candidate pool (Ministral-14B, Qwen3-Coder-30B, Devstral-123B, GPT-OSS-20B, GPT-OSS-120B, Qwen3-235B).
BARP achieves the highest Acc@1 under full observability (52.69\%), MLP achieves the highest Acc@1-L (39.16\%), Naive Bayes achieves the highest DCG (0.4512), and Hybrid-LLM achieves the highest L-DCG (0.4151).
Several methods exhibit policy collapse under partial observability: BARP, Avenger, Graph-Router, and Hybrid-LLM all suffer sharp drops in Acc@1-L (below 23\%), while methods like Naive Bayes and Emb-FT remain comparatively stable.

\begin{table*}[t]
\centering
\renewcommand{\arraystretch}{1.3}
\resizebox{1.0\textwidth}{!}{%
\begin{tabular}{l l c c c c c c c c}
\hline
& & \multicolumn{4}{c}{\textbf{Fully-Observable}} & \multicolumn{4}{c}{\textbf{Partially-Observable}} \\
\cline{3-6} \cline{7-10}
\textbf{Method} & \textbf{Base Model}
& \textbf{Acc@1 (\%)} & \textbf{Acc@1-L (\%)} & \textbf{DCG} & \textbf{L-DCG}
& \textbf{Acc@1 (\%)} & \textbf{Acc@1-L (\%)} & \textbf{DCG} & \textbf{L-DCG} \\
\hline
\rowcolor{gray!15} \multicolumn{10}{c}{\textit{Naive Baselines}} \\
\rowcolor{blue!15}
Oracle (Perfect) & --
& 100.00 & 100.00 & 1.0000 & 1.0000
& 100.00 & 100.00 & 1.0000 & 1.0000 \\
\rowcolor{blue!6}
Oracle (Best-of-N) & --
& 82.67 & 70.65 & 0.7791 & 0.7521
& 94.89 & 77.35 & 0.8781 & 0.8313 \\
Random & --
& 41.64 & 16.99 & 0.3010 & 0.2211
& 41.64 & 16.99 & 0.3010 & 0.2211 \\
Route-To-Devstral & --
& 44.41 & 21.20 & 0.3519 & 0.2674
& 44.41 & 21.20 & 0.3519 & 0.2674 \\
Route-To-Ministral & --
& 22.93 & 22.79 & 0.2286 & 0.2284
& 22.93 & 22.79 & 0.2286 & 0.2284 \\
Route-To-OSS-20B & --
& 39.36 & 10.77 & 0.2611 & 0.1694
& 39.36 & 10.77 & 0.2611 & 0.1694 \\
Route-To-OSS-120B & --
& 48.96 & 14.64 & 0.3281 & 0.2181
& 48.96 & 14.64 & 0.3281 & 0.2181 \\
Route-To-Qwen3-Coder & --
& 29.56 & 17.33 & 0.2441 & 0.2130
& 29.56 & 17.33 & 0.2441 & 0.2130 \\
Route-To-Qwen3-235B & --
& 60.36 & 13.26 & 0.3652 & 0.1967
& 60.36 & 13.26 & 0.3652 & 0.1967 \\
\rowcolor{gray!15} \multicolumn{10}{c}{\textit{Representation-based Routers}} \\
KNN & Qwen3-Emb-0.6B
& 48.20 & 38.40 & 0.4414 & 0.4118
& 51.59 & 21.82 & 0.3774 & 0.2814 \\
Naive Bayes & Qwen3-Emb-0.6B
& \underline{51.73} & 36.12 & \textbf{0.4512} & 0.4021
& 50.69 & \textbf{31.15} & \textbf{0.4225} & \textbf{0.3680} \\
MLP & Qwen3-Emb-0.6B
& 45.23 & \textbf{39.16} & 0.4282 & \underline{0.4122}
& \textbf{58.70} & 20.17 & 0.4014 & 0.2687 \\
Emb-FT & Qwen3-Emb-0.6B
& 44.89 & 36.95 & 0.4170 & 0.3925
& 55.94 & \underline{23.48} & \underline{0.4095} & \underline{0.3045} \\
Hybrid-LLM & DeBERTa Large
& 46.48 & \underline{39.02} & 0.4351 & \textbf{0.4151}
& \underline{58.08} & 13.88 & 0.3595 & 0.2049 \\
Graph-Router & Qwen3-Emb-0.6B
& 46.13 & 13.47 & 0.3086 & 0.2016
& 53.87 & 13.40 & 0.3426 & 0.2012 \\
BARP & Qwen3-Emb-0.6B
& \textbf{52.69} & 32.73 & \underline{0.4434} & 0.3739
& 22.93 & 22.79 & 0.2286 & 0.2284 \\
Avenger & Qwen3-Emb-0.6B
& 42.54 & 36.12 & 0.3990 & 0.3848
& 43.92 & 11.67 & 0.2873 & 0.1790 \\
\hline
\end{tabular}%
}
\caption{Routing performance on \textsc{RW-Agentic} (6-model pool) under fully- and partially-observable settings.
\textbf{Bold} = best; \underline{underline} = second best.}
\label{tab:router_comparison_7way}
\end{table*}

\subsubsection{3-Model Pool}

Table~\ref{tab:router_comparison_agentic} presents results with the original 3-model pool (Qwen3-235B, GPT-OSS-20B, Nemotron-9B).
Orange-shaded cells indicate collapsed routing policies that always select a single model; these are excluded from best/second-best comparisons.

\begin{table*}[t]
\centering
\renewcommand{\arraystretch}{1.3}
\resizebox{1.0\textwidth}{!}{%
\begin{tabular}{l l c c c c c c c c}
\hline
& & \multicolumn{4}{c}{\textbf{Fully-Observable}} & \multicolumn{4}{c}{\textbf{Partially-Observable}} \\
\cline{3-6} \cline{7-10}
\textbf{Method} & \textbf{Base Model}
& \textbf{Acc@1 (\%)} & \textbf{Acc@1-L (\%)} & \textbf{DCG} & \textbf{L-DCG}
& \textbf{Acc@1 (\%)} & \textbf{Acc@1-L (\%)} & \textbf{DCG} & \textbf{L-DCG} \\
\hline
\rowcolor{gray!15} \multicolumn{10}{c}{\textit{Naive Baselines}} \\
\rowcolor{blue!15}
Oracle (Perfect) & -- & 100.00 & 100.00 & 1.0000 & 1.0000 & 100.00 & 100.00 & 1.0000 & 1.0000 \\
\rowcolor{blue!6}
Oracle (Best-of-8) & -- & 86.81 & 73.46 & 0.8176 & 0.7931 & 88.16 & 75.19 & 0.8325 & 0.8051 \\
Random & -- & 41.89 & 26.38 & 0.3589 & 0.3161 & 41.89 & 26.38 & 0.3589 & 0.3161 \\
Route-To-Qwen & -- & 67.95 & 38.92 & 0.5706 & 0.4971 & 67.95 & 38.92 & 0.5706 & 0.4971 \\
Route-To-Nemotron & -- & 43.03 & 20.54 & 0.3376 & 0.2615 & 43.03 & 20.54 & 0.3376 & 0.2615 \\
Route-To-OSS & -- & 51.19 & 40.54 & 0.4698 & 0.4462 & 51.19 & 40.54 & 0.4698 & 0.4462 \\
\rowcolor{gray!15} \multicolumn{10}{c}{\textit{Generative AI Routers}} \\
Zero-Shot Prompt & Qwen-2.5-0.5B-ins
& \cellcolor{orange!18}67.95 & \cellcolor{orange!18}38.92 & \cellcolor{orange!18}0.5706 & \cellcolor{orange!18}0.4971
& \cellcolor{orange!18}67.95 & \cellcolor{orange!18}38.92 & \cellcolor{orange!18}0.5706 & \cellcolor{orange!18}0.4971 \\
Gen-SFT & Qwen-2.5-0.5B-ins
& \underline{76.43} & \underline{58.32} & \underline{0.6954} & \underline{0.6581}
& \underline{75.62} & \underline{50.59} & \underline{0.6622} & \underline{0.6022} \\
Zero-Shot Prompt & Qwen-2.5-1.5B-ins
& \cellcolor{orange!18}67.95 & \cellcolor{orange!18}38.92 & \cellcolor{orange!18}0.5706 & \cellcolor{orange!18}0.4971
& \cellcolor{orange!18}67.95 & \cellcolor{orange!18}38.92 & \cellcolor{orange!18}0.5706 & \cellcolor{orange!18}0.4971 \\
Gen-SFT & Qwen-2.5-1.5B-ins
& 76.16 & 58.32 & 0.6935 & 0.6576
& 74.76 & 50.05 & 0.6546 & 0.5953 \\
\rowcolor{gray!15} \multicolumn{10}{c}{\textit{Representation-based Routers}} \\
KNN & Qwen3-Embedding-0.6B
& 75.73 & 57.84 & 0.6889 & 0.6512
& 72.54 & 48.76 & 0.6341 & 0.5742 \\
MLP & Qwen3-Embedding-0.6B
& 77.41 & 59.24 & 0.7050 & 0.6673
& 77.35 & 55.46 & 0.6907 & 0.6411 \\
Emb-FT & Modern-base-0.1B
& 77.78 & 58.16 & 0.7034 & 0.6604
& \cellcolor{orange!18}67.95 & \cellcolor{orange!18}38.92 & \cellcolor{orange!18}0.5706 & \cellcolor{orange!18}0.4971 \\
Emb-FT & Qwen3-Embedding-0.6B
& \underline{78.05} & \textbf{60.00} & \textbf{0.7118} & \textbf{0.6747}
& \textbf{78.54} & \textbf{57.35} & \textbf{0.7053} & \textbf{0.6583} \\
Hybrid-LLM & DeBERTa Large
& \textbf{78.16} & \underline{59.35} & \underline{0.7104} & \underline{0.6706}
& \underline{78.05} & \underline{56.16} & \underline{0.6980} & \underline{0.6479} \\
Graph-Router & Qwen3-Embedding-0.6B
& \cellcolor{orange!18}67.95 & \cellcolor{orange!18}38.97 & \cellcolor{orange!18}0.5708 & \cellcolor{orange!18}0.4974
& \cellcolor{orange!18}67.95 & \cellcolor{orange!18}38.92 & \cellcolor{orange!18}0.5706 & \cellcolor{orange!18}0.4971 \\
BARP & Qwen3-Embedding-0.6B
& 75.89 & 52.16 & 0.6696 & 0.6137
& 68.22 & 39.41 & 0.5741 & 0.5013 \\
Avenger & Qwen3-Embedding-0.6B
& 75.08 & 58.43 & 0.6877 & 0.6529
& 69.84 & 44.38 & 0.6021 & 0.5388 \\
\hline
\end{tabular}%
}
\caption{Routing performance on \textsc{RW-Agentic} under fully- and partially-observable settings.
Oracle (Perfect) and Oracle (Best-of-$N$) denote theoretical and empirical upper bounds, respectively.
\colorbox{orange!18}{Orange cells} indicate collapsed routing (always selects a single model) and are excluded from best/second-best comparisons.
\textbf{Bold} = best; \underline{underline} = second best among non-collapsed methods.}
\label{tab:router_comparison_agentic}
\end{table*}

\section{Query Encoder Scaling}\label{app:encoder-scaling}

\begin{table}[t]
\centering
\resizebox{\linewidth}{!}{%
\begin{tabular}{l|ccc|ccc}
\toprule
& \multicolumn{3}{c|}{\textbf{Fully-Obs}} & \multicolumn{3}{c}{\textbf{Partially-Obs}} \\
& 0.6B & 8B & $\Delta$ & 0.6B & 8B & $\Delta$ \\
\midrule
KNN     & .397 & .405 & \gc{+0.8} & .222 & .223 & \gc{+0.2} \\
MLP     & .430 & .439 & \gc{+1.0} & .216 & .259 & \gc{+4.3} \\
BaRP    & .298 & .162 & \rc{-13.6} & .157 & .200 & \gc{+4.4} \\
Avenger & .371 & .379 & \gc{+0.8} & .149 & .149 & \gc{+0.0} \\
\bottomrule
\end{tabular}%
}
\caption{L-DCG: 0.6B vs.\ 8B encoder ($\Delta$ in pp).}
\label{tab:encoder-scale}
\end{table}

We compare \texttt{Qwen3-Embedding-0.6B} (1024-dim) against
\texttt{Qwen3-Embedding-8B} (4096-dim) across four methods that support
both encoders. Table~\ref{tab:encoder-scale} reports the results under
fully and partially-observable settings.

Scaling the encoder from 0.6B to 8B ($13\times$ parameters, $4\times$
embedding dimension) yields surprisingly modest gains. Under full
observability, KNN and Avenger improve by less than 1 pp in L-DCG, and MLP
gains only 1 pp. BaRP \emph{degrades} sharply, likely
because its lightweight decision head (64-dim) cannot effectively leverage
the 4096-dim representations, leading to mode collapse.

Under partial observability, the picture is mixed: MLP and BaRP benefit
modestly from the larger encoder ($+4$ pp each), but KNN and Avenger show
near-zero improvement. Notably, Avenger with 8B produces identical L-DCG
to 0.6B ($0.149$) under cold-start, which both collapse.

These results suggest that \textbf{routing quality is bottlenecked by the
training signal, not encoder capacity}. A $13\times$ parameter increase
provides at most $+1$ pp L-DCG under full observability and $+4$ pp under
partial observability, while warm-up initialization (Figure~\ref{fig:warmup-wave}, see Appendix~\ref{app:additional-figures})
recovers $+15$--$18$ pp with the same 0.6B encoder. Investing in better
training data (e.g., diverse warm-up labels) is far more effective than
scaling the encoder.

\section{RouterWorld Prompt Details}
\label{sec:appendix-rw-prompts}

This section documents all prompts used in the RouterWorld data collection pipeline and routing methods.
Figures~\ref{fig:prompt-rollout-math}--\ref{fig:prompt-rollout-qa} show the prompts used to roll out candidate models on core evaluation tasks.
Figures~\ref{fig:prompt-rollout-alfworld}--\ref{fig:prompt-rollout-workarena} show the agent rollout prompts.
Figures~\ref{fig:prompt-judge-code}--\ref{fig:prompt-judge-chat} present the LLM-as-Judge prompts for automated correctness annotation.
Figure~\ref{fig:prompt-user-query} shows the user query annotation prompt.
Figures~\ref{fig:prompt-router}--\ref{fig:prompt-router-rag} present the generative router prompts used for SFT training and inference.

\subsection{Core Task Rollout Prompts}
\label{sec:appendix-rollout-core}

Each benchmark category uses a standardized prompt template to query candidate models during data collection.

\begin{figure}[!t]
\centering
\begin{tcolorbox}[
  colback=white, colframe=promptframe, coltitle=white,
  fonttitle=\bfseries\sffamily\small,
  title={Math Rollout Prompt~~{\normalfont\color{white!80}\scriptsize MATH500 / AIME / FinQA / LiveMathBench}},
  attach boxed title to top left={yshift=-2mm, xshift=4mm},
  boxed title style={colback=catRollout, rounded corners, sharp corners=south},
  rounded corners, arc=3pt, boxrule=0.5pt, width=\linewidth,
  top=4pt, bottom=4pt, left=6pt, right=6pt,
]
\begin{tcolorbox}[
  colback=promptsysbg, colframe=promptsys!30,
  rounded corners, arc=2pt, boxrule=0.3pt,
  top=2pt, bottom=2pt, left=4pt, right=4pt,
]
\footnotesize\ttfamily
Solve the following math problem step by step. The last line of your response should only contain your final answer inside a \textbackslash boxed\{\} command.\\[3pt]
\textcolor{promptvar}{\{Question\}}\\[3pt]
Remember to put your answer inside \textbackslash boxed\{\}.
\end{tcolorbox}
\end{tcolorbox}
\caption{Rollout prompt for mathematical reasoning benchmarks.}
\label{fig:prompt-rollout-math}
\end{figure}

\begin{figure}[!t]
\centering
\begin{tcolorbox}[
  colback=white, colframe=promptframe, coltitle=white,
  fonttitle=\bfseries\sffamily\small,
  title={MCQ Rollout Prompt~~{\normalfont\color{white!80}\scriptsize MMLUPro / GPQA / TruthfulQA / MedQA / BrainTeaser}},
  attach boxed title to top left={yshift=-2mm, xshift=4mm},
  boxed title style={colback=catRollout, rounded corners, sharp corners=south},
  rounded corners, arc=3pt, boxrule=0.5pt, width=\linewidth,
  top=4pt, bottom=4pt, left=6pt, right=6pt,
]
\begin{tcolorbox}[
  colback=promptsysbg, colframe=promptsys!30,
  rounded corners, arc=2pt, boxrule=0.3pt,
  top=2pt, bottom=2pt, left=4pt, right=4pt,
]
\footnotesize\ttfamily
Answer the following \textcolor{promptvar}{\{subject\}} question. The last line of your response should be of the following format: `Answer: \$LETTER' (without quotes) where LETTER is one of \textcolor{promptvar}{\{candidates\}}.\\[3pt]
\textcolor{promptvar}{\{question\}}\\[3pt]
\textcolor{promptvar}{\{options\}}
\end{tcolorbox}
{\scriptsize\color{textmuted} The \texttt{\{subject\}} field is populated for MMLUPro; other MCQ benchmarks omit it.}
\end{tcolorbox}
\caption{Rollout prompt for multiple-choice benchmarks.}
\label{fig:prompt-rollout-mcq}
\end{figure}

\begin{figure}[!t]
\centering
\begin{tcolorbox}[
  colback=white, colframe=promptframe, coltitle=white,
  fonttitle=\bfseries\sffamily\small,
  title={Code Rollout Prompt~~{\normalfont\color{white!80}\scriptsize HumanEval / MBPP / StudentEval / LiveCodeBench}},
  attach boxed title to top left={yshift=-2mm, xshift=4mm},
  boxed title style={colback=catRollout, rounded corners, sharp corners=south},
  rounded corners, arc=3pt, boxrule=0.5pt, width=\linewidth,
  top=4pt, bottom=4pt, left=6pt, right=6pt,
]
\begin{tcolorbox}[
  colback=promptsysbg, colframe=promptsys!30,
  rounded corners, arc=2pt, boxrule=0.3pt,
  top=2pt, bottom=2pt, left=4pt, right=4pt,
]
\footnotesize\ttfamily
You are an expert Python programmer, and here is your task:\\
\textcolor{promptvar}{\{question\}}\\[3pt]
Your code should pass these tests:\\
\textcolor{promptvar}{\{tests\}}\\[3pt]
Your code should start with a markdown code block.
\end{tcolorbox}
\end{tcolorbox}
\caption{Rollout prompt for code generation benchmarks.}
\label{fig:prompt-rollout-code}
\end{figure}

\begin{figure}[!t]
\centering
\begin{tcolorbox}[
  colback=white, colframe=promptframe, coltitle=white,
  fonttitle=\bfseries\sffamily\small,
  title={Logical Reasoning Rollout Prompt~~{\normalfont\color{white!80}\scriptsize BBH (3-shot CoT)}},
  attach boxed title to top left={yshift=-2mm, xshift=4mm},
  boxed title style={colback=catRollout, rounded corners, sharp corners=south},
  rounded corners, arc=3pt, boxrule=0.5pt, width=\linewidth,
  top=4pt, bottom=4pt, left=6pt, right=6pt,
]
\begin{tcolorbox}[
  colback=promptsysbg, colframe=promptsys!30,
  rounded corners, arc=2pt, boxrule=0.3pt,
  top=2pt, bottom=2pt, left=4pt, right=4pt,
]
\footnotesize\ttfamily
Answer the logical question, Write a line of the form ``Answer: \$ANSWER'' at the end of your response. Let's think step by step.\\[3pt]
Here are 3 examples to help you understand how to answer the question:\\
\textcolor{promptvar}{\{template\}}\\[3pt]
Question: \textcolor{promptvar}{\{question\}}
\end{tcolorbox}
\end{tcolorbox}
\caption{Rollout prompt for BBH logical reasoning (3-shot with chain-of-thought).}
\label{fig:prompt-rollout-bbh}
\end{figure}

\begin{figure}[h]
\centering
\begin{tcolorbox}[
  colback=white, colframe=promptframe, coltitle=white,
  fonttitle=\bfseries\sffamily\small,
  title={Deductive Reasoning Rollout Prompt~~{\normalfont\color{white!80}\scriptsize Knights \& Knaves}},
  attach boxed title to top left={yshift=-2mm, xshift=4mm},
  boxed title style={colback=catRollout, rounded corners, sharp corners=south},
  rounded corners, arc=3pt, boxrule=0.5pt, width=\linewidth,
  top=4pt, bottom=4pt, left=6pt, right=6pt,
]
\begin{tcolorbox}[
  colback=promptsysbg, colframe=promptsys!30,
  rounded corners, arc=2pt, boxrule=0.3pt,
  top=2pt, bottom=2pt, left=4pt, right=4pt,
]
\footnotesize\ttfamily
Your task is to solve a logical reasoning problem. You are given set of statements from which you must logically deduce the identity of a set of characters.\\[3pt]
You must infer the identity of each character. First, explain your reasoning. At the end of your answer, you must clearly state the identity of each character by following the format:\\[3pt]
CONCLUSION:\\
(1) ...\\
(2) ...\\
(3) ...\\[3pt]
\#\#\# Question:\\
\textcolor{promptvar}{\{question\}}
\end{tcolorbox}
\end{tcolorbox}
\caption{Rollout prompt for Knights \& Knaves deductive reasoning.}
\label{fig:prompt-rollout-kk}
\end{figure}

\begin{figure}[h]
\centering
\begin{tcolorbox}[
  colback=white, colframe=promptframe, coltitle=white,
  fonttitle=\bfseries\sffamily\small,
  title={Commonsense Rollout Prompt~~{\normalfont\color{white!80}\scriptsize Winogrande}},
  attach boxed title to top left={yshift=-2mm, xshift=4mm},
  boxed title style={colback=catRollout, rounded corners, sharp corners=south},
  rounded corners, arc=3pt, boxrule=0.5pt, width=\linewidth,
  top=4pt, bottom=4pt, left=6pt, right=6pt,
]
\begin{tcolorbox}[
  colback=promptsysbg, colframe=promptsys!30,
  rounded corners, arc=2pt, boxrule=0.3pt,
  top=2pt, bottom=2pt, left=4pt, right=4pt,
]
\footnotesize\ttfamily
You are given a *cloze* sentence containing a blank marked by underscores `\_\_\_'. Only \textbf{one} of the two options correctly fills the blank while preserving commonsense.\\[3pt]
After your reasoning, end with a line \textbf{exactly} in the form: `Answer: \$LETTER', where `LETTER' is \textbf{A} or \textbf{B}.\\[3pt]
Sentence: \textcolor{promptvar}{\{sentence\}}\\
A) \textcolor{promptvar}{\{option1\}}~~B) \textcolor{promptvar}{\{option2\}}
\end{tcolorbox}
\end{tcolorbox}
\caption{Rollout prompt for Winogrande commonsense reasoning.}
\label{fig:prompt-rollout-wino}
\end{figure}

\begin{figure}[!t]
\centering
\begin{tcolorbox}[
  colback=white, colframe=promptframe, coltitle=white,
  fonttitle=\bfseries\sffamily\small,
  title={ALFWorld ReAct Agent Prompt~~{\normalfont\color{white!80}\scriptsize 6 task types $\times$ 3 few-shot examples}},
  attach boxed title to top left={yshift=-2mm, xshift=4mm},
  boxed title style={colback=catAgent, rounded corners, sharp corners=south},
  rounded corners, arc=3pt, boxrule=0.5pt, width=\linewidth,
  top=4pt, bottom=4pt, left=6pt, right=6pt,
]
{\footnotesize The ALFWorld agent uses ReAct-style few-shot demonstrations. For each of the 6 task types (\texttt{put}, \texttt{clean}, \texttt{heat}, \texttt{cool}, \texttt{puttwo}, \texttt{examine}), 3 in-context examples show the think--act--observe loop. Two prompt variants exist: original and a fixed version with corrected action verbs.}

\vspace{3pt}
\begin{tcolorbox}[
  colback=promptsysbg, colframe=promptsys!30,
  fonttitle=\footnotesize\sffamily\bfseries, coltitle=promptsys,
  title={Example: \texttt{react\_put\_0} (abbreviated)},
  rounded corners, arc=2pt, boxrule=0.3pt,
  top=2pt, bottom=2pt, left=4pt, right=4pt,
]
\footnotesize\ttfamily
You are in the middle of a room. Looking quickly around you, you see a cabinet 4, a cabinet 3, ...\\
Your task is to: put some spraybottle on toilet.\\
> think: To solve the task, I need to find and take a spraybottle, then put it on toilet.\\
OK.\\
> think: First I need to find a spraybottle. A spraybottle is more likely to appear in cabinet (1-4), countertop (1), ...\\
OK.\\
> go to cabinet 2\\
The cabinet 2 is closed.\\
> open cabinet 2\\
You open the cabinet 2. In it, you see a candle 1, and a spraybottle 2.\\
> take spraybottle 2 from cabinet 2\\
You pick up the spraybottle 2 from the cabinet 2.\\
> go to toilet 1\\
> put spraybottle 2 in/on toilet 1\\
You put the spraybottle 2 in/on the toilet 1.
\end{tcolorbox}
\end{tcolorbox}
\caption{ALFWorld agent rollout prompt with ReAct few-shot demonstrations.}
\label{fig:prompt-rollout-alfworld}
\end{figure}

\begin{figure}[h]
\centering
\begin{tcolorbox}[
  colback=white, colframe=promptframe, coltitle=white,
  fonttitle=\bfseries\sffamily\small,
  title={Emotion Recognition Rollout Prompt~~{\normalfont\color{white!80}\scriptsize EmoryNLP / MELD / DailyDialog}},
  attach boxed title to top left={yshift=-2mm, xshift=4mm},
  boxed title style={colback=catRollout, rounded corners, sharp corners=south},
  rounded corners, arc=3pt, boxrule=0.5pt, width=\linewidth,
  top=4pt, bottom=4pt, left=6pt, right=6pt,
]
\begin{tcolorbox}[
  colback=promptsysbg, colframe=promptsys!30,
  rounded corners, arc=2pt, boxrule=0.3pt,
  top=2pt, bottom=2pt, left=4pt, right=4pt,
]
\footnotesize\ttfamily
Given a conversation history and a current utterance, follow these steps to identify the emotion of the current utterance from the given options. The emotion should be determined based on both the conversation context and the current utterance.\\
The last line of your response should be of the following format: `Answer: \$LETTER' (without quotes) where LETTER is one of ABCDEFG. Let's think step by step.\\[3pt]
\textcolor{promptvar}{\{conversation\_history\}}\\[3pt]
Current utterance: \textcolor{promptvar}{\{utterance\}}\\
Options: \textcolor{promptvar}{\{options\}}
\end{tcolorbox}
\end{tcolorbox}
\caption{Rollout prompt for emotion recognition (shared across EmoryNLP, MELD, DailyDialog).}
\label{fig:prompt-rollout-emotion}
\end{figure}

\begin{figure}[h]
\centering
\begin{tcolorbox}[
  colback=white, colframe=promptframe, coltitle=white,
  fonttitle=\bfseries\sffamily\small,
  title={Open QA Rollout Prompt~~{\normalfont\color{white!80}\scriptsize SimpleQA}},
  attach boxed title to top left={yshift=-2mm, xshift=4mm},
  boxed title style={colback=catRollout, rounded corners, sharp corners=south},
  rounded corners, arc=3pt, boxrule=0.5pt, width=\linewidth,
  top=4pt, bottom=4pt, left=6pt, right=6pt,
]
\begin{tcolorbox}[
  colback=promptsysbg, colframe=promptsys!30,
  rounded corners, arc=2pt, boxrule=0.3pt,
  top=2pt, bottom=2pt, left=4pt, right=4pt,
]
\footnotesize\ttfamily
Answer the following question.\\
Question: \textcolor{promptvar}{\{question\}}\\[3pt]
Answer:
\end{tcolorbox}
\end{tcolorbox}
\caption{Rollout prompt for open-ended factual QA.}
\label{fig:prompt-rollout-qa}
\end{figure}

\subsection{Agentic Task Rollout Prompts}
\label{sec:appendix-rollout-agent}

For agentic benchmarks, each candidate model is deployed as an interactive agent producing valid actions in a structured format. Unlike core tasks where a single prompt--response pair suffices, agentic rollouts involve multi-turn interactions between the model and the environment. Notably, several candidate models required extensive prompt engineering (${\sim}$2 hours per model) to reliably extract valid actions in the expected format, as models frequently produced verbose explanations or malformed action strings instead of the required structured output.

\begin{figure}[h]
\centering
\begin{tcolorbox}[
  colback=white, colframe=promptframe, coltitle=white,
  fonttitle=\bfseries\sffamily\small,
  title={WorkArena Web Agent System Prompt~~{\normalfont\color{white!80}\scriptsize ServiceNow tasks}},
  attach boxed title to top left={yshift=-2mm, xshift=4mm},
  boxed title style={colback=catAgent, rounded corners, sharp corners=south},
  rounded corners, arc=3pt, boxrule=0.5pt, width=\linewidth,
  top=4pt, bottom=4pt, left=6pt, right=6pt,
]
\begin{tcolorbox}[
  colback=promptsysbg, colframe=promptsys!30,
  rounded corners, arc=2pt, boxrule=0.3pt,
  top=2pt, bottom=2pt, left=4pt, right=4pt,
]
\footnotesize\ttfamily
You are an autonomous web agent interacting with a ServiceNow instance.\\[3pt]
Your task is to complete knowledge work tasks by interacting with the web browser.\\[3pt]
Available actions:\\
- click(element\_id): Click on an element by its ID\\
- fill(element\_id, text): Fill a text field with the given text\\
- select(element\_id, option): Select an option from a dropdown\\
- scroll(direction): Scroll the page up or down\\
- goto(url): Navigate to a specific URL\\
- send\_msg\_to\_user(message): Send a message to the user\\[3pt]
Output format: \texttt{```action} \texttt{<your\_action\_here>} \texttt{```}\\[3pt]
Rules:\\
1. Output exactly ONE action per response\\
2. Always use the action code block format\\
3. Be precise with element IDs from the observation\\
4. Follow the task instructions carefully
\end{tcolorbox}
\end{tcolorbox}
\caption{WorkArena web agent system prompt for ServiceNow knowledge-work tasks.}
\label{fig:prompt-rollout-workarena}
\end{figure}

\subsection{LLM-as-Judge Prompts}
\label{sec:appendix-judge}

After collecting model responses, we use LLM-as-Judge prompts to automatically annotate correctness. Each task category has a dedicated judge prompt that outputs its verdict inside \texttt{<a>...</a>} tags.

\begin{figure}[h]
\centering
\begin{tcolorbox}[
  colback=white, colframe=promptframe, coltitle=white,
  fonttitle=\bfseries\sffamily\small,
  title={LLM-as-Judge: Code Generation},
  attach boxed title to top left={yshift=-2mm, xshift=4mm},
  boxed title style={colback=catJudge, rounded corners, sharp corners=south},
  rounded corners, arc=3pt, boxrule=0.5pt, width=\linewidth,
  top=4pt, bottom=4pt, left=6pt, right=6pt,
]
\begin{tcolorbox}[
  colback=promptsysbg, colframe=promptsys!30,
  rounded corners, arc=2pt, boxrule=0.3pt,
  top=2pt, bottom=2pt, left=4pt, right=4pt,
]
\footnotesize\ttfamily
You are an expert evaluator tasked with judging the correctness of code generation responses.\\[3pt]
User Instruction/Problem: \textcolor{promptvar}{[[prompt]]}\\
Model's Generated Solution: \textcolor{promptvar}{[[response]]}\\[3pt]
Save your evaluation inside <a></a>, such as <a>correct</a> or <a>incorrect</a>.\\[3pt]
Instructions:\\
1. Extract the code from the model's response (ignore explanatory text)\\
2. Evaluate if the code correctly solves the problem\\
3. Check for logical errors, bugs, or hallucinated functionality\\
4. Ignore minor stylistic issues if logic is correct\\
5. Consider the solution correct if it would produce the right output\\
6. Partially correct solutions that solve the core problem can be considered correct\\
7. Consider it incorrect if they hardcode\\
8. Don't repeat the full code; quick judge whether the code is acceptable
\end{tcolorbox}
\end{tcolorbox}
\caption{LLM-as-Judge prompt for code generation tasks.}
\label{fig:prompt-judge-code}
\end{figure}

\begin{figure}[h]
\centering
\begin{tcolorbox}[
  colback=white, colframe=promptframe, coltitle=white,
  fonttitle=\bfseries\sffamily\small,
  title={LLM-as-Judge: Open-Ended QA},
  attach boxed title to top left={yshift=-2mm, xshift=4mm},
  boxed title style={colback=catJudge, rounded corners, sharp corners=south},
  rounded corners, arc=3pt, boxrule=0.5pt, width=\linewidth,
  top=4pt, bottom=4pt, left=6pt, right=6pt,
]
\begin{tcolorbox}[
  colback=promptsysbg, colframe=promptsys!30,
  rounded corners, arc=2pt, boxrule=0.3pt,
  top=2pt, bottom=2pt, left=4pt, right=4pt,
]
\footnotesize\ttfamily
You are an expert evaluator tasked with judging the correctness of question responses.\\[3pt]
Given:\\
- A ground truth answer: \textcolor{promptvar}{[[ground\_truth]]}\\
- A model's response: \textcolor{promptvar}{[[response]]}\\[3pt]
Instructions:\\
1. Extract the final answer (True/False, Yes/No, Entailment/Neutral/Contradiction, etc.)\\
2. Ignore reasoning; only extract the final answer\\
3. Normalize both answers (handle case differences, common variations)\\
4. For semantic equivalence: answers that mean the same thing should be correct\\
5. Respond with exactly one word: ``correct'' or ``incorrect''
\end{tcolorbox}
\end{tcolorbox}
\caption{LLM-as-Judge prompt for open-ended QA tasks.}
\label{fig:prompt-judge-qa}
\end{figure}

\begin{figure}[h]
\centering
\begin{tcolorbox}[
  colback=white, colframe=promptframe, coltitle=white,
  fonttitle=\bfseries\sffamily\small,
  title={LLM-as-Judge: Multiple-Choice},
  attach boxed title to top left={yshift=-2mm, xshift=4mm},
  boxed title style={colback=catJudge, rounded corners, sharp corners=south},
  rounded corners, arc=3pt, boxrule=0.5pt, width=\linewidth,
  top=4pt, bottom=4pt, left=6pt, right=6pt,
]
\begin{tcolorbox}[
  colback=promptsysbg, colframe=promptsys!30,
  rounded corners, arc=2pt, boxrule=0.3pt,
  top=2pt, bottom=2pt, left=4pt, right=4pt,
]
\footnotesize\ttfamily
You are an expert evaluator tasked with judging the correctness of multiple-choice question responses.\\[3pt]
Given:\\
- A ground truth answer: \textcolor{promptvar}{[[ground\_truth]]}\\
- A model's response: \textcolor{promptvar}{[[response]]}\\[3pt]
Instructions:\\
1. Extract the final answer (look for patterns like A, B, C, D, E, F)\\
2. Some answers may contain reasoning. Ignore it; only extract the final answer\\
3. Compare it with the ground truth answer\\
4. Respond with exactly one word: ``correct'' or ``incorrect''
\end{tcolorbox}
\end{tcolorbox}
\caption{LLM-as-Judge prompt for multiple-choice tasks.}
\label{fig:prompt-judge-mcq}
\end{figure}

\begin{figure}[h]
\centering
\begin{tcolorbox}[
  colback=white, colframe=promptframe, coltitle=white,
  fonttitle=\bfseries\sffamily\small,
  title={LLM-as-Judge: SQL},
  attach boxed title to top left={yshift=-2mm, xshift=4mm},
  boxed title style={colback=catJudge, rounded corners, sharp corners=south},
  rounded corners, arc=3pt, boxrule=0.5pt, width=\linewidth,
  top=4pt, bottom=4pt, left=6pt, right=6pt,
]
\begin{tcolorbox}[
  colback=promptsysbg, colframe=promptsys!30,
  rounded corners, arc=2pt, boxrule=0.3pt,
  top=2pt, bottom=2pt, left=4pt, right=4pt,
]
\footnotesize\ttfamily
You are an expert evaluator tasked with judging the correctness of SQL query responses.\\[3pt]
User Instruction: \textcolor{promptvar}{[[prompt]]}\\[3pt]
Given:\\
- A ground truth SQL query: \textcolor{promptvar}{[[ground\_truth]]}\\
- A model's generated SQL query: \textcolor{promptvar}{[[response]]}\\[3pt]
Instructions:\\
1. Extract the SQL query from the model's response (ignore explanatory text)\\
2. Ensure the table and column names are not hallucinated\\
3. You can ignore DISTINCT, or ROUND such details\\
4. Compare the logical equivalence of both queries\\
5. Consider queries correct if semantically equivalent, even if syntactically different\\
6. Partially correct can also be considered as correct
\end{tcolorbox}
\end{tcolorbox}
\caption{LLM-as-Judge prompt for SQL generation tasks (relaxed variant).}
\label{fig:prompt-judge-sql}
\end{figure}

\begin{figure}[h]
\centering
\begin{tcolorbox}[
  colback=white, colframe=promptframe, coltitle=white,
  fonttitle=\bfseries\sffamily\small,
  title={LLM-as-Judge: Tool Use / Function Calling},
  attach boxed title to top left={yshift=-2mm, xshift=4mm},
  boxed title style={colback=catJudge, rounded corners, sharp corners=south},
  rounded corners, arc=3pt, boxrule=0.5pt, width=\linewidth,
  top=4pt, bottom=4pt, left=6pt, right=6pt,
]
\begin{tcolorbox}[
  colback=promptsysbg, colframe=promptsys!30,
  rounded corners, arc=2pt, boxrule=0.3pt,
  top=2pt, bottom=2pt, left=4pt, right=4pt,
]
\footnotesize\ttfamily
You are an expert evaluator tasked with judging the correctness of tool use and function calling responses.\\[3pt]
Given:\\
- A ground truth tool call: \textcolor{promptvar}{[[ground\_truth]]}\\
- A model's response: \textcolor{promptvar}{[[response]]}\\[3pt]
Instructions:\\
1. Extract the tool/function call (look for function names, parameters, arguments)\\
2. Verify: correct tool name, all required parameters present, values match, types correct\\
3. Ignore reasoning text---focus only on the actual tool call\\
4. Minor formatting differences are acceptable if semantic meaning is identical\\
5. Count it correct if they use function name such as ``Currency Finder'' not ``currency.find''\\
6. Don't have to be over-strict (not case-sensitive, not bool-sensitive)
\end{tcolorbox}
\end{tcolorbox}
\caption{LLM-as-Judge prompt for tool use / function calling tasks.}
\label{fig:prompt-judge-tool}
\end{figure}

\begin{figure}[h]
\centering
\begin{tcolorbox}[
  colback=white, colframe=promptframe, coltitle=white,
  fonttitle=\bfseries\sffamily\small,
  title={LLM-as-Judge: Personal Chat},
  attach boxed title to top left={yshift=-2mm, xshift=4mm},
  boxed title style={colback=catJudge, rounded corners, sharp corners=south},
  rounded corners, arc=3pt, boxrule=0.5pt, width=\linewidth,
  top=4pt, bottom=4pt, left=6pt, right=6pt,
]
\begin{tcolorbox}[
  colback=promptsysbg, colframe=promptsys!30,
  rounded corners, arc=2pt, boxrule=0.3pt,
  top=2pt, bottom=2pt, left=4pt, right=4pt,
]
\footnotesize\ttfamily
You are an expert evaluator tasked with judging the quality and appropriateness of chatbot responses in personal conversations.\\[3pt]
Given:\\
- A conversation context: \textcolor{promptvar}{[[prompt]]}\\
- A model's response: \textcolor{promptvar}{[[response]]}\\[3pt]
Instructions:\\
1. Evaluate if the response is contextually relevant\\
2. Check if the tone is suitable for personal/casual conversation\\
3. Assess if the response maintains appropriate boundaries\\
4. Verify no harmful, offensive, or inappropriate content\\
5. Consider if the response moves the conversation forward constructively\\[3pt]
``good'' if: addresses context naturally, suitable tone, helpful, maintains boundaries.\\
``bad'' if: ignores context, overly formal/robotic, unhelpful, inappropriate content.\\[3pt]
Don't be too strict since NL allows for rooms. Only brief reasons (\textbf{one sentence}) then evaluation.
\end{tcolorbox}
{\scriptsize\color{textmuted} Uses a ``good/bad'' scale instead of ``correct/incorrect''.}
\end{tcolorbox}
\caption{LLM-as-Judge prompt for personal chat tasks.}
\label{fig:prompt-judge-chat}
\end{figure}

\subsection{User Query Annotation Prompt}
\label{sec:appendix-user-query}

To make the routing dataset realistic, we convert formal benchmark task descriptions into natural user queries. The agent pipeline variant additionally samples from 15 diverse rephrasing styles (e.g., ``direct command'', ``polite request'', ``search keywords'', ``Show me...'', ``Help me pick...'') to increase linguistic diversity.

\begin{figure}[h]
\centering
\begin{tcolorbox}[
  colback=white, colframe=promptframe, coltitle=white,
  fonttitle=\bfseries\sffamily\small,
  title={User Query Annotation Prompt},
  attach boxed title to top left={yshift=-2mm, xshift=4mm},
  boxed title style={colback=catAnnot, rounded corners, sharp corners=south},
  rounded corners, arc=3pt, boxrule=0.5pt, width=\linewidth,
  top=4pt, bottom=4pt, left=6pt, right=6pt,
]
\begin{tcolorbox}[
  colback=promptsysbg, colframe=promptsys!30,
  rounded corners, arc=2pt, boxrule=0.3pt,
  top=2pt, bottom=2pt, left=4pt, right=4pt,
]
\footnotesize\ttfamily
You are simulating a human USER interacting with an AI assistant. Generate a natural, colloquial query that a real user would ask for the given task.\\[3pt]
\# Input\\
You will receive a formal task prompt that may include:\\
- User requests and requirements\\
- Referenced data or context\\
- Success criteria or expected outcomes\\[3pt]
\# Guidelines\\
Transform the formal prompt into natural, colloquial language by:\\
- Using casual phrasing (as a real user would speak)\\
- Omitting references to ``tabular data'' or backend systems\\
- For objective answer tasks (multiple choice, yes/no):\\
\quad - Avoid formal answer formats\\
\quad - Integrate answer choices naturally into the question\\
- Removing any meta-instructions or formatting requirements\\
- Keeping the core intent and requirements clear\\
- Don't always start with ``Hey, I'm ...'', natural and concise please\\[3pt]
\# Output Format\\
Return ONLY the converted user query wrapped in tags: <a>[Your natural query here]</a>\\[3pt]
\# Complete Task Prompt\\
\textcolor{promptvar}{[[prompt]]}
\end{tcolorbox}
\end{tcolorbox}
\caption{User query annotation prompt. Converts formal benchmark descriptions into natural conversational-form queries for the routing dataset.}
\label{fig:prompt-user-query}
\end{figure}

\subsection{Generative Router Prompts}
\label{sec:appendix-router-prompt}

The generative routing approach uses the following prompt templates for both SFT training data construction and inference. The model zoo section is dynamically populated with candidate model names and their measured latency statistics (TTFT, inter-token latency). The expected output is a single model index wrapped in \texttt{<answer>...</answer>} tags.

\begin{figure}[h]
\centering
\begin{tcolorbox}[
  colback=white, colframe=promptframe, coltitle=white,
  fonttitle=\bfseries\sffamily\small,
  title={Generative Router Prompt~~{\normalfont\color{white!80}\scriptsize SFT training \& inference}},
  attach boxed title to top left={yshift=-2mm, xshift=4mm},
  boxed title style={colback=catRouter, rounded corners, sharp corners=south},
  rounded corners, arc=3pt, boxrule=0.5pt, width=\linewidth,
  top=4pt, bottom=4pt, left=6pt, right=6pt,
]
\begin{tcolorbox}[
  colback=promptsysbg, colframe=promptsys!30,
  rounded corners, arc=2pt, boxrule=0.3pt,
  top=2pt, bottom=2pt, left=4pt, right=4pt,
]
\footnotesize\ttfamily
You are an expert LLM router. Given a user query, you need to select the best model that can:\\
1. Answer the query correctly\\
2. Respond as fast as possible\\[3pt]
Available Models:\\
Index 0: Mistral-7B-Instruct-v0.3 (TTFT: 100.0ms, ITL: 0.46ms)\\
Index 1: Qwen3-235B-A22B-Instruct-2507 (TTFT: 2258.0ms, ITL: 2.70ms)\\
{\color{textmuted}\itshape ...remaining models...}\\[3pt]
Given the following query, predict which model is most likely to answer correctly AND be the fastest among correct models.\\[3pt]
Query: \textcolor{promptvar}{\{query\}}\\[3pt]
Instructions:\\
- Consider model capabilities and latency\\
- Output ONLY the model index number (0-\textcolor{promptvar}{\{max\_model\_idx\}}) that you predict will be correct and fastest\\
- Format your answer as: <answer>MODEL\_INDEX</answer>\\[3pt]
Your prediction:
\end{tcolorbox}
\vspace{3pt}
\begin{tcolorbox}[
  colback=promptoutbg, colframe=promptout!30,
  fonttitle=\footnotesize\sffamily\bfseries, coltitle=promptout,
  title={Training Target},
  rounded corners, arc=2pt, boxrule=0.3pt,
  top=2pt, bottom=2pt, left=4pt, right=4pt,
]
\footnotesize\ttfamily <answer>\textcolor{promptvar}{\{label\}}</answer>
\quad{\normalfont\color{textmuted}\scriptsize where \texttt{label} = index of the fastest correct model}
\end{tcolorbox}
\end{tcolorbox}
\caption{Generative router prompt used for SFT training and inference. The model zoo section is populated dynamically with measured latency statistics.}
\label{fig:prompt-router}
\end{figure}

\newpage
\pagebreak

\begin{figure}[!t]
\centering
\begin{tcolorbox}[
  colback=white, colframe=promptframe, coltitle=white,
  fonttitle=\bfseries\sffamily\small,
  title={RAG-Augmented Router Prompt},
  attach boxed title to top left={yshift=-2mm, xshift=4mm},
  boxed title style={colback=catRouter, rounded corners, sharp corners=south},
  rounded corners, arc=3pt, boxrule=0.5pt, width=\linewidth,
  top=4pt, bottom=4pt, left=6pt, right=6pt,
]
\begin{tcolorbox}[
  colback=promptsysbg, colframe=promptsys!30,
  rounded corners, arc=2pt, boxrule=0.3pt,
  top=2pt, bottom=2pt, left=4pt, right=4pt,
]
\footnotesize\ttfamily
You are an expert LLM router. Given a user query, select the best model that answers correctly and fastest.\\[3pt]
\textcolor{promptvar}{\{model\_zoo\}}\\[3pt]
Here are similar queries from past routing decisions (correct model shown):\\
~~Example 1 [MATH500]: ``Solve for x in 3x+7=22'' $\to$ Model 4 (deepseek-math, 342ms)\\
~~Example 2 [MMLUPro]: ``Which element has atomic number 79?'' $\to$ Model 0 (Mistral-7B, 187ms)\\
~~{\color{textmuted}\itshape ...}\\[3pt]
Now route this new query:\\
Query: \textcolor{promptvar}{\{query\}}\\[3pt]
Based on the similar examples above, output ONLY the model index (0-\textcolor{promptvar}{\{max\_model\_idx\}}) that will answer correctly and fastest.\\
Answer: <answer>
\end{tcolorbox}
\end{tcolorbox}
\caption{RAG-augmented router prompt. Retrieved similar queries from the training set are included as few-shot examples to improve routing accuracy.}
\label{fig:prompt-router-rag}
\end{figure}

\end{document}